\documentclass[lettersize,journal]{IEEEtran}
\usepackage{amsmath,amsfonts}
\usepackage{algorithmic}
\usepackage{array}
\usepackage[caption=false,font=normalsize,labelfont=sf,textfont=sf]{subfig}
\usepackage{textcomp}
\usepackage{stfloats}
\usepackage{url}
\usepackage{verbatim}
\usepackage{graphicx}
\usepackage{cite}
\usepackage{algorithm}
\usepackage{booktabs}
\usepackage{xcolor}
\usepackage{amssymb} 
\usepackage{colortbl}
\definecolor{softblue}{HTML}{0070c0}
\definecolor{upgreen}{HTML}{24c042}
\definecolor{mygray}{HTML}{d9d9d9}

\begin{document}

\title{SChanger: Change Detection from a Semantic Change  and Spatial Consistency  Perspective}

\author{Ziyu Zhou, Keyan Hu, Yutian Fang, Xiaoping Rui*
\thanks{Ziyu Zhou, Yutian Fang, and Xiaoping Rui (corresponding author) are
with the School of Earth Sciences and Engineering, Hohai University, Nanjing 211100, China
(e-mail: ziyuzhou@hhu.edu.cn; yutianfang@hhu.edu.cn; ruixp@hhu.edu.cn).}
\thanks{Keyan Hu is with the School of Geosciences and Info-physics, Central South University, Changsha 410100, China
(e-mail: phycheor@gmail.com).}}

\markboth{IEEE JOURNAL OF SELECTED TOPICS IN APPLIED EARTH OBSERVATIONS AND REMOTE SENSING}%
{Shell \MakeLowercase{\textit{et al.}}: A Sample Article Using IEEEtran.cls for IEEE Journals}

\maketitle

\begin{abstract}
  Change detection is a key task in Earth observation applications. Recently, deep learning methods have demonstrated strong performance and widespread application. However, change detection faces data scarcity due to the labor-intensive process of accurately aligning remote sensing images of the same area, which limits the performance of deep learning algorithms. To address the data scarcity issue, we develop a fine-tuning strategy called the Semantic Change Network (SCN). We initially pre-train the model on single-temporal supervised tasks to acquire prior knowledge of instance feature extraction. The model then employs a shared-weight Siamese architecture and extended Temporal Fusion Module (TFM) to preserve this prior knowledge and is fine-tuned on change detection tasks. The learned semantics for identifying all instances is changed to focus on identifying only the changes. Meanwhile, we observe that the locations of changes between the two images are spatially identical, a concept we refer to as spatial consistency. We introduce this inductive bias through an attention map that is generated by large-kernel convolutions and applied to the features from both time points. This enhances the modeling of multi-scale changes and helps capture underlying relationships in change detection semantics. We develop a binary change detection model utilizing these two strategies. The model is validated against state-of-the-art methods on six datasets, surpassing all benchmark methods and achieving F1 scores of 92.87\%, 86.43\%, 68.95\%, 97.62\%, 84.58\%, and 93.20\% on the LEVIR-CD, LEVIR-CD+, S2Looking, CDD, SYSU-CD, and WHU-CD datasets, respectively.

\end{abstract}


\section{Introduction}
\begin{figure*}[!t]
  \centering
  \includegraphics[width=7in]{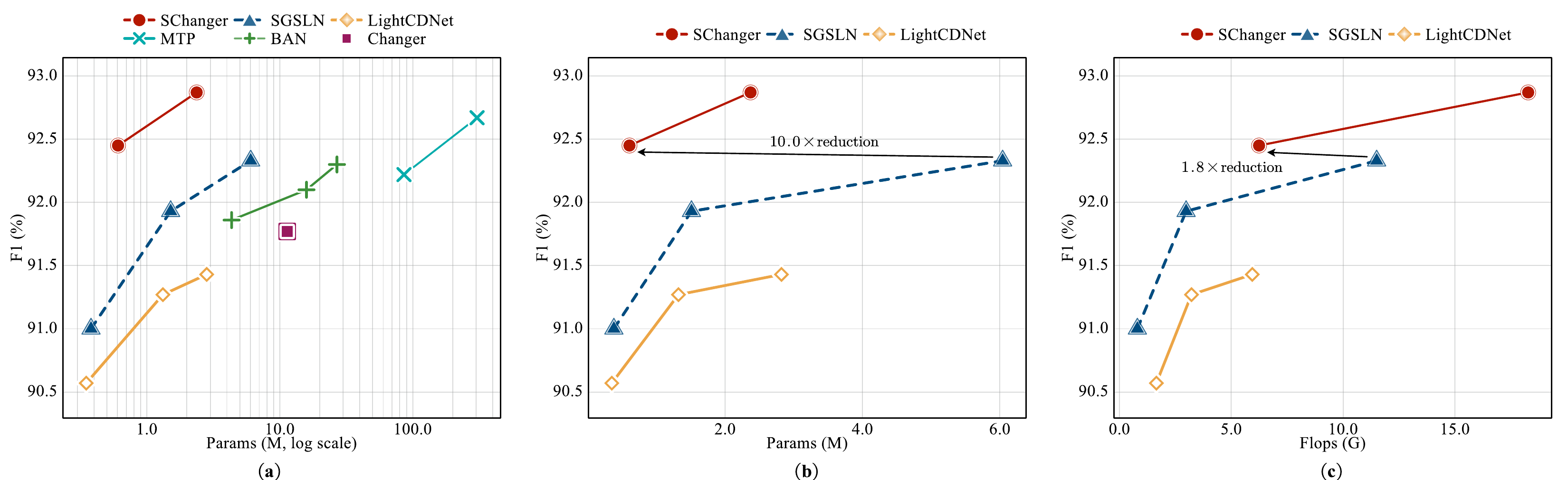}
  \caption{Params/Flops vs. Performance. All performance results are obtained using a single model, with Flops calculated based on a tensor of shape 1$\times$2$\times$256$\times$256. SChanger achieves significant improvements in efficiency regarding Params and Flops, while delivering comparable or superior performance on the LEVIR-CD dataset relative to previous CD models.}
  \label{fig-l_compare}
\end{figure*}
\IEEEPARstart{C}{hange} detection (CD) aims to identify alterations 
on the Earth's surface by analyzing multi-temporal remote sensing images captured at different time points over the same geo-graphical area~\cite{singh1989review}. 
This technique is widely utilized in urban development monitoring~\cite{wen2021change}, disaster assessment~\cite{zheng2021building}, and land use planning~\cite{zhu2022land}. 
High-resolution remote sensing imagery has made remote sensing change detection (RSCD) possible, providing the clarity and detail necessary for 
accurate and precise analysis of changes.

With the emergence of large-scale remote sensing datasets, deep learning technology tailored for RSCD has rapidly advanced. Fully convolutional Siamese networks~\cite{daudt2018fully} are widely recognized as a pioneering deep learning approach for CD, marking the beginning of extensive deep learning applications in this field. Current methods in CD encompass a variety of approaches, including convolutional neural networks (CNNs)-based techniques (e.g., Changer~\cite{fang2023changer}, SGSLN~\cite{zhao2023exchanging}), transformer-based approaches (e.g., ChangeFormer~\cite{bandara2022transformer}, BiT~\cite{chen2021remote}), and state space model-based methods (e.g., RSM-CD~\cite{zhao2024rs}). However, the prerequisite for training robust and high-performing models is the availability of high-quality annotated datasets.

Acquiring and accurately aligning remote sensing images of the same region for CD tasks often requires considerable time and effort. The registration process may introduce errors due to differences in sensors or varying environmental conditions, which in turn limits the availability of usable image pairs~\cite{chen2020spatial,lebedev2018change}. This data scarcity often results in a performance bottleneck for deep learning algorithms~\cite{long2021toward}. The size of CD datasets is much smaller than that of single-temporal datasets like the WHU Building Dataset~\cite{ji2018fully}, the Inria Aerial Image Labeling Dataset~\cite{maggiori2017can}, and general image datasets like ImageNet~\cite{russakovsky2015imagenet} and MS COCO~\cite{lin2014microsoft}. The discrepancy in dataset sizes between single-temporal and multi-temporal tasks directly affects the effectiveness of model training in CD. Pre-training on single-temporal datasets is a viable solution. However, integrating these pre-trained weights into models with dual-temporal image inputs is a key challenge. During this process, it is also critical to retain as much pre-trained knowledge as possible.

This study introduces SChanger, a new family of CD models designed for high-precision predictions with efficient parameters and computational complexity. To tackle data scarcity, we pre-train the Semantic Prior Network (SPNet) on single-temporal segmentation tasks, incorporating prior knowledge of instances. For dual-temporal inputs in CD tasks, we propose a fine-tuning method called the Semantic Change Network (SCN). It uses a shared-weight Siamese network architecture and the Temporal Fusion Module (TFM) to retain and adapt the knowledge that was learned before. The Siamese network aligns dual-temporal features in a consistent semantic space, while the TFM resolves channel mismatches. By fine-tuning on CD tasks, the semantics of the features shift from the single-temporal segmentation domain to the dual-temporal CD domain. The Spatial Consistency Attention Module (SCAM) in the model adds an inductive bias of spatial consistency. This effectively combines dual-temporal information and focuses feature extraction on areas where building changes happen. The model also includes a Lightweight Feature Enhancement Module (LFTM) for feature enhancement and a Multi-Scale Fusion Segmentation Head (MSFSH) for multi-scale information output.

We extensively evaluate SChanger on six popular CD datasets. The result for the LEVIR-CD dataset is shown in Fig.~\ref{fig-l_compare}(a), which indicates that SChanger outperforms previous lightweight models such as LightCDNet~\cite{xing2023lightcdnet} and SGSLN~\cite{zhao2023exchanging}. More importantly, compared to previous leading lightweight models, SChanger achieves a 10.0$\times$ reduction in parameters, as shown in Fig.~\ref{fig-l_compare}(b), and a 1.8$\times$ reduction in floating point operations (Flops), as shown in Fig.~\ref{fig-l_compare}(c), while maintaining similar performance. We summarize our contributions as follows:
\begin{enumerate} 
  \item SCN is introduced as a fine-tuning strategy for CD tasks, utilizing a shared-weight Siamese network and strategically positioned TFM modules. This design adjusts the model's computational logic to enhance accuracy.

  \item SCAM is designed to incorporate a shared attention map derived from large-kernel convolutions on features from both time points, introducing an inductive bias of spatial consistency. This mechanism enhances the model's ability to capture long-range dependencies and infer relationships within change semantics.
  
  \item The proposed model, SChanger, achieves notable efficiency improvements by reducing parameter count and Flops while outperforming previous state-of-the-art models in CD tasks and maintaining competitive accuracy.
\end{enumerate}

\section{Related Works}
\subsection{Pre-training and Fine-tuning Paradigm for RSCD}
In recent years, the pre-training and fine-tuning transfer learning paradigm has been widely adopted in RSCD research~\cite{fang2023changer,zhang2020deeply}. Pre-training is the process of training the model on large-scale datasets to learn general feature representations and patterns. When task-specific datasets are limited, pre-training can significantly enhance the model's generalization and performance. Fine-tuning builds upon the pre-trained model by further training on specific datasets, allowing the model to adapt to specific tasks or domains while reducing the high costs and risks of overfitting~\cite{pan2009survey}.

However, traditional pre-training datasets such as ImageNet lack essential domain-specific features such as buildings, roads, and vegetation, which are critical for remote sensing tasks. This limitation impedes the effective transfer of pre-trained weights to these applications. Recent studies show that pre-training models from scratch on remote sensing datasets (RSP) significantly enhances performance in remote sensing tasks. For supervised pre-training, Wang et al.~\cite{wang2022empirical} trained architectures like ResNet~\cite{he2016deep} and Swin Transformer~\cite{liu2021swin} on the Million-AID dataset, demonstrating that RSP models outperform those using ImageNet (IMP). Similarly, Bastani et al.~\cite{bastani2023satlaspretrain} showed that pre-training on the larger SATLAS dataset yielded superior results. In addition, Wang et al.~\cite{wang2024mtp} also achieved notable gains using multi-task pre-training on the SAMRS dataset. The larger dataset size and more detailed supervision contribute to better performance. For unsupervised pre-training, Cong et al.~\cite{cong2022satmae} applied the masked autoencoders (MAE) technique on the fMoW-Sentinel dataset, improving transfer learning performance across several downstream tasks. For single-temporal supervised learning, Zheng et al.~\cite{zheng2021change} used XOR operations to generate CD image pairs from two unpaired labeled images. Simultaneously, Zheng et al~\cite{zheng2024changen2} introduced a resolution-scalable diffusion transformer that can generate time-series images and their corresponding semantics. We can conclude that the alignment between the visual domain of pre-training and the domain of the target task tends to positively influence the model's performance.

While pre-training can significantly boost model performance, fine-tuning, particularly on small datasets, often causes overfitting, limiting the model's overall effectiveness. Most transfer learning approaches assume fixed model capacity, applying fine-tuning to either the whole model or just the task-specific output layer. Recent methods have focused on expanding model capacity to better align with downstream tasks~\cite{wang2017growing,rebuffi2017learning}. Sung et al.~\cite{sung2022lst} introduced the Ladder Side-Tuning (LST) method, which enhances adaptability by freezing the base model and adding a side network for training. Li et al.~\cite{li2024new} introduced the BAN method, an LST extension tailored for CD tasks, optimizing dual-temporal image analysis. Although LST methods enhance a model's adaptability, they introduce too many extra parameters, leading to inefficiencies. Therefore, a key challenge is finding ways to maintain parameter efficiency while adapting to the dual-image input required for CD tasks.

Unlike data-centric approaches in single-temporal supervised learning~\cite{zheng2021change,zheng2024single}, where the model structure for pretraining and fine-tuning remains unchanged and only the data organization is modified, our work introduces a novel method, SCN, which adopts a model-centric perspective for transformation. By leveraging the TFM and Siamese network to expand the model's capacity for handling dual-temporal inputs and modifying the attention mechanism's computational logic during the fine-tuning phase, SCN effectively transforms single-temporal models into dual-temporal ones. This enables the utilization of pre-trained weights from single-temporal data, resulting in higher performance.

\subsection{Deep Learning Models in RSCD}
With the rapid advancements in remote sensing technologies and artificial intelligence, deep learning techniques have swiftly found broad applications in remote sensing~\cite{wang2018scene,zhai2021hyperspectral}. By extracting higher-level feature representations of image data through multi-level processing~\cite{varghese2018changenet}, deep learning has significantly propelled the development of RSCD technology~\cite{peng2019end}. Common CD architectures include single-branch and dual-branch (Siamese) networks~\cite{zhang2021hdfnet}. Single-branch networks combine two images from different times before the encoder to make processing of the inputs easier. Siamese networks, on the other hand, use a dual encoder or decoder structure with shared weights so that each image can be processed separately for better feature learning.

Typically, single-branch networks achieve image fusion by concatenating or applying absolute differencing to the input data. However, this often introduces redundancy, obscuring critical temporal variations and hindering effective extraction of meaningful change features. To address this, Papadomanolaki et al.~\cite{papadomanolaki2019detecting} incorporated long short-term memory (LSTM) into CNNs. Xing et al.~\cite{xing2023lightcdnet} used channel attention mechanisms. Zheng et al.~\cite{zheng2021clnet} integrated multi-scale features and multi-level semantic context. However, combining the semantic features of two images too early may not provide adequate guidance for modeling changes in features, often leading to lower accuracy.
In contrast, Dual-branch networks maintain separate feature extraction paths for each temporal image, providing clearer insights into changes over time. FC-Siam-conc and FC-Siam-diff networks become foundational for many subsequent CD frameworks~\cite{daudt2018fully}. Chen et al.~\cite{chen2022fccdn} state that combining dual-temporal features may introduce irrelevant background information, degrading performance. To preserve object-level differences and enhance interaction, Zhang et al.~\cite{zhang2020deeply} combined original and differential features with channel and spatial attention mechanisms. Fang et al.~\cite{fang2023changer} further improved dual-temporal interaction by exchanging feature channels between the two time points. These methods show how important it is to improve feature interaction in Siamese networks, which do better at CD tasks than single-branch networks because they can handle inputs more independently.

Multi-scale instances in remote sensing images challenge a model's ability to accurately detect and represent structures. Extending the model's receptive field is a common solution. Bandara and Patel~\cite{bandara2022transformer} introduced the Vision Transformer (ViT) to improve long-sequence modeling. Despite ViT's strong performance in natural image tasks~\cite{liu2021swin,dosovitskiy2020image}, its quadratic complexity makes it expensive for high-resolution remote sensing. Furthermore, ViTs require far more training data than CNNs~\cite{bao2021beit}, limiting their benefits in CD tasks, especially with limited data.
Depthwise convolutions
To emulate ViT's ability to capture long-range dependencies, CNN models have increased convolutional kernel sizes. However, this leads to a significant rise in parameters and computational complexity. To address this, Yu and Koltun~\cite{yu2015multi} introduced dilated convolutions, expanding the receptive field without adding many parameters. Zhang et al.\cite{zhang2018triplet} further enhanced multi-scale feature extraction with parallel dilated convolutions. , known for their low parameter count, have also gained popularity for efficiently achieving large kernels. These methods have improved performance, as seen in models like ConvNeXt\cite{liu2022convnet} and RepLKNet~\cite{ding2022scaling}. CNNs remain the dominant choice for RSCD data processing, efficiently extracting feature information while maintaining parameter and computational efficiency.

Thus, this study adopts convolutional methods, with the network's overall architecture utilizing a fully Siamese network to more independently extract features from each temporal instance. The introduction of SCAM ensures effective fusion of bi-temporal information. SCAM improves its accuracy in detecting changes by effectively distinguishing between altered and unaltered instances using large kernel attention~\cite{guo2023visual}.
\section{Method}
\begin{figure*}[!t]
  \centering
  \includegraphics[width=6in]{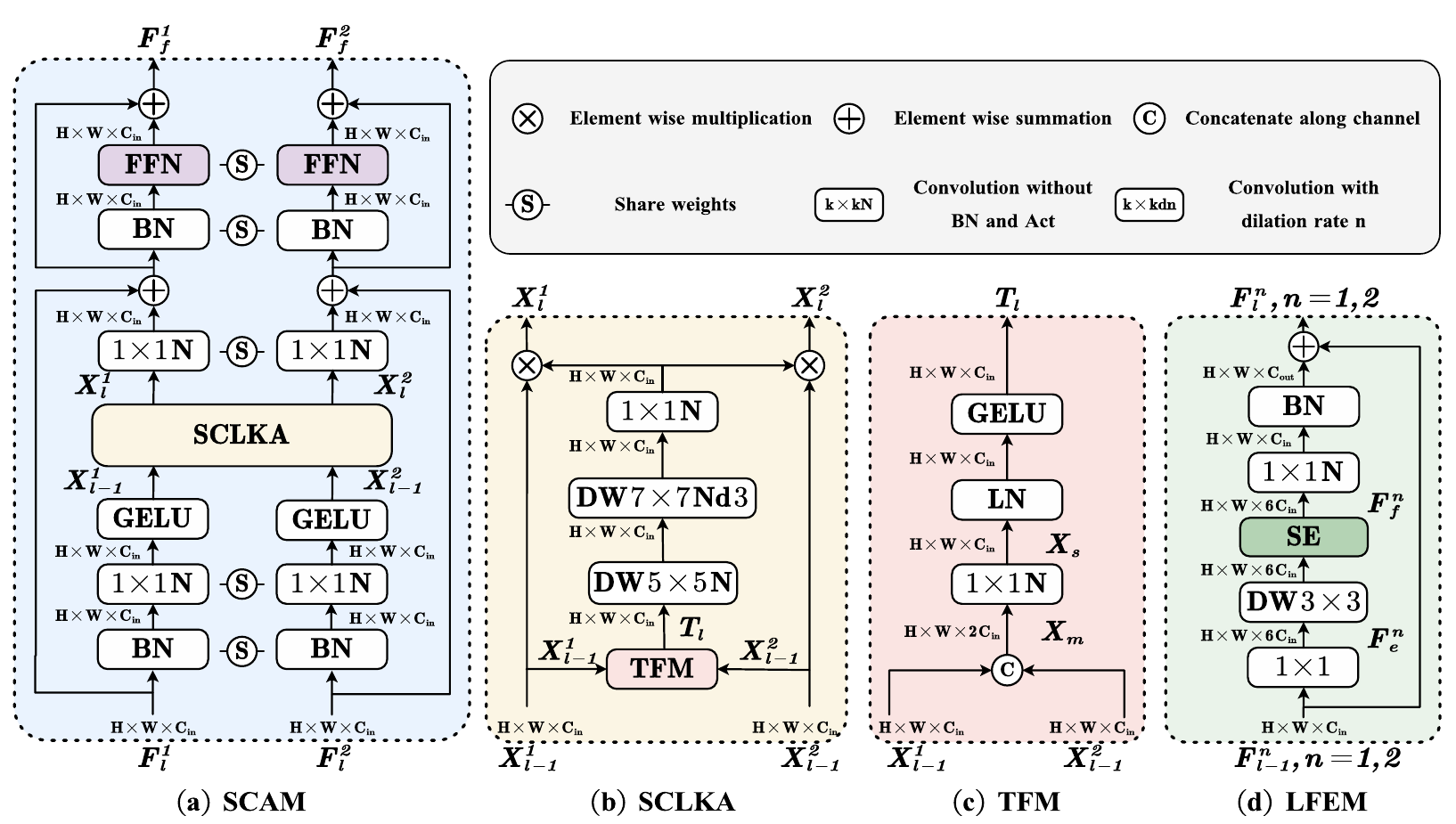}
  \caption{Overview of the module design. (a) illustrates each SCAM structure (Section~\ref{subsec:SCAM}) that facilitates dual-temporal information interaction. (b) illustrates each SCLKA Structure (Section~\ref{subsec:SCAM}). (c) illustrates each TFM structure (Section~\ref{subsec:TFM}), where dual-temporal features are fused. (d) illustrates each LFEM structure (Section~\ref{subsec:LFEM}), where dual-temporal features are initially projected into a similar feature space.}
  \label{fig_1}
\end{figure*}

In this section, we explore the modular design and overall structure of the network. First, we provide an overview of the SCAM, TFM, LFEM, and MSFSH modules, outlining their functions and contributions to the network architecture. We then explain how SPNet is built for single-temporal supervised pretraining. Next, we demonstrate how the SCN strategy enhances SPNet's capacity to handle temporal changes, resulting in the SCchanger model, an effective network for CD. Finally, we discuss the training process and key technical details of the network.
\subsection{Spatial Consistency Attention Module}\label{subsec:SCAM}
In CD tasks, effectively capturing dual-temporal information requires incorporating inductive biases into the model. Traditional Siamese networks, which process features independently from each temporal state, fail to account for the relationships and changes between them. To address this limitation, we propose SCAM as illustrated in Fig.~\ref{fig_1}(a). SCAM improves the model's ability to detect changes by introducing an inherent inductive bias that is specific to CD. SCAM is composed of two main components: Spatial Consistency Large Kernel Attention (SCLKA) and the TFM. The details of SCAM and SCLKA are explained here, while TFM is covered in Section~\ref{subsec:TFM}.

To enhance the model's ability to capture long-range dependencies, we combine attention mechanism with large-kernel convolution. We employ a large-kernel convolution decomposition technique~\cite{guo2023visual}, which splits large-kernel convolutions into smaller, sequential operations. It improves the model's capacity to generate more precise attention maps, reducing the number of parameters and computational complexity. As shown in Fig.~\ref{fig_1}(b), the SCLKA module generates its output using the following steps:
\begin{equation}
  \label{T_l}
  T_l=\text{TFM}\left( X_{l-1}^{1},X_{l-1}^{2} \right) .
  \end{equation}
\begin{equation}
    \label{S_l}
    S_l=DWConv_{k_2\times k_2,d}\left( DWConv_{k_1\times k_1}\left( T_l \right) \right).
\end{equation}
\begin{equation}
  \label{Attn}
  Attn=Conv_{1\times 1}\left( S_l \right).
\end{equation}
Where~$T_l\in R^{H\times W\times C_{in}}$ represents the dual-temporal features fused by the TFM module, and~$S_l\in R^{H\times W\times C_{in}}$ denotes the spatially fused features obtained through depthwise and dilated convolutions. $Attn\in R^{H\times W\times C_{in}}$ refers to the attention map generated through channel interaction using pointwise convolutions. $k_1$ and $k_2$ represent the kernel sizes of the first and second convolutions, respectively, while $d$ denotes the dilation rate of the second convolution.
For consistency and to facilitate a more meaningful comparison, the same parameter configuration as in LKA \cite{guo2023visual} is used. Specifically, $k_1=5$, $k_2=7$, and $d=3$ are set to approximate a $21 \times 21$ convolution, aiming to achieve a balance between performance and computational efficiency.
\begin{figure}[!t]
  \centering
  \includegraphics[width=2.5in]{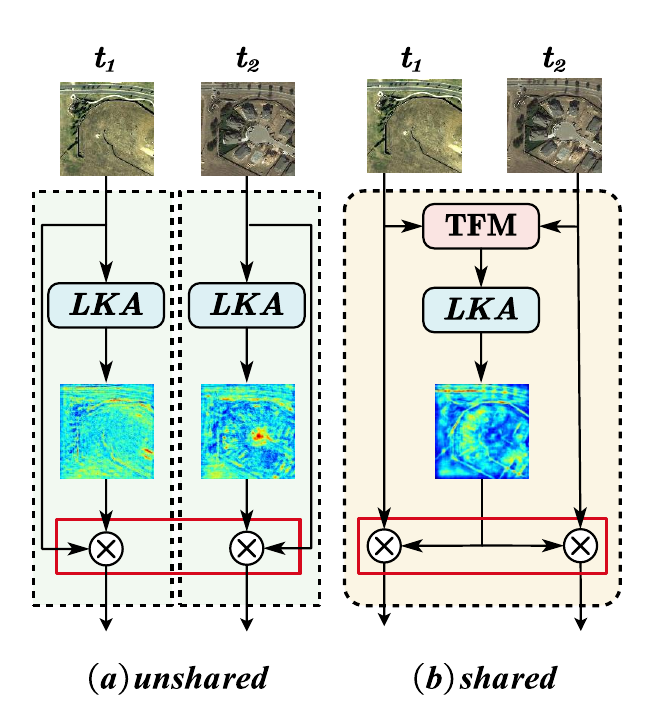}
  \caption{Illustration of Shared Attention Mechanism for Dual-Temporal Images in CD. (a) Unshared Attention, (b) Shared Attention.}
  \label{fig-shared}
\end{figure}
In previous methods~\cite{guo2023visual}, attention mechanisms are applied independently to the dual-temporal images, as shown in Fig.~\ref{fig-shared}(a). Due to the nature of CD tasks, the segmented instances exist only in a single image, which leads to wasted computations for the changing instances in the other image. This also makes it more prone to introducing noise.
To address this issue, An inductive bias, referred to as spatial consistency, is introduced into the model to capture changes between dual-temporal images. This bias assumes that regions experiencing changes (such as transitions from bare land to buildings) will retain their semantic identity over time. Essentially, if a region is identified as changing at one time point, it should still be recognized as a "changing" region at another time point, even though the land cover types may differ. As a result, the same "changing" attention map can be applied to both feature sets. As shown in Fig.~\ref{fig-shared}(b), a shared attention map is used to influence both time-point images.
Mathematically, this is expressed as:
\begin{equation}
  \label{X_l^1}
  X_{l}^{1}=Attn\otimes X_{l-1}^{1}.
\end{equation}
\begin{equation}
  \label{X_l^2}
  X_{l}^{2}=Attn\otimes X_{l-1}^{2}.
\end{equation}
Where, $X_{l-1}^{1},X_{l-1}^{2}\in R^{H\times W\times C_{in}}$ denote the input feature maps for the two temporal images, with $\otimes$ representing element-wise multiplication. This approach enables the model to capture broader contextual information, effectively integrating CD signals while maintaining the fidelity of dual-temporal feature representations.

Finally, within the SCAM structure, the features $\mathrm{F}_{\mathrm{l}}^{1}$ and $\mathrm{F}_{\mathrm{l}}^{2}$ of the l-th layer are successively passed through Batch Normalization (BN), a $1\times 1$ convolution, the GELU activation function, the SCLKA module, 
and a Feed Forward Network to extract feature representations. Additionally, we adopt a weight-sharing Siamese network structure to ensure consistent feature extraction across both temporal inputs.
\begin{figure*}[!t]
  \centering
  \includegraphics[width=6.5in]{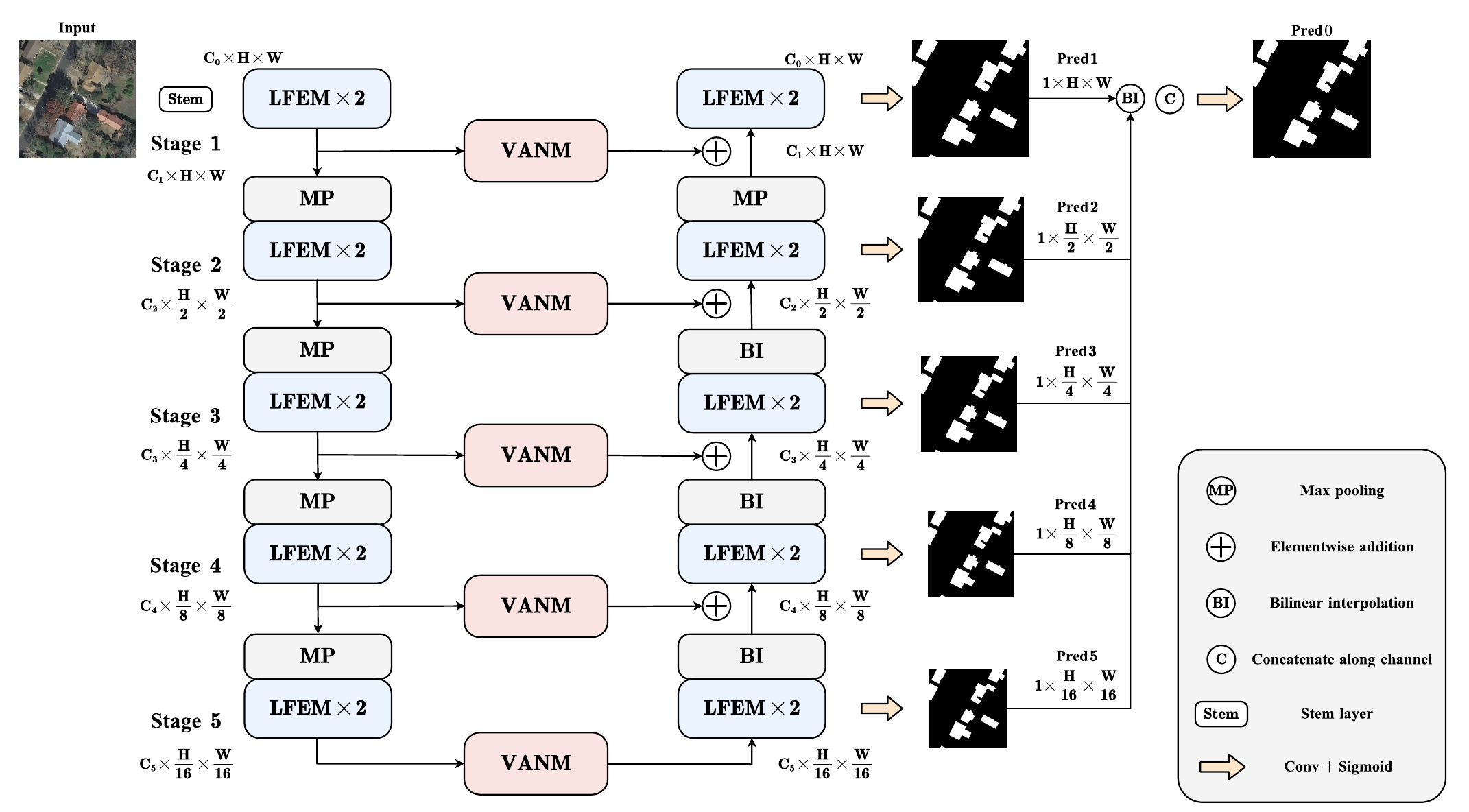}
  \caption{Overall Structure of SPNet. LFEM denotes the Lightweight Feature Enhancement Module (Section~\ref{subsec:LFEM}), while VANM denotes the Visual Attention Network Module~\cite{guo2023visual}.}
  \label{fig-BPnet}
\end{figure*}
\subsection{Temporal Fusion Module}\label{subsec:TFM}
Feature fusion in CD models often relies on simple or convolutional strategies. Simple methods like addition, subtraction, and concatenation are vulnerable to noise, hindering accurate CD~\cite{daudt2018fully,fang2023changer}. Convolutional techniques attempt to address this by fusing dual-temporal features via channel concatenation, followed by convolution to improve fusion efficiency~\cite{zhao2023exchanging}. However, the normalization process after convolution still requires further discussion.
Normalization techniques like BN often overlook the unique temporal characteristics of dual-temporal features. Applying BN directly can cause inconsistencies between time steps, leading to training instability and fluctuations~\cite{shen2020powernorm}. To mitigate this, Layer Normalization (LN)~\cite{ba2016layer} is used to independently normalize each sample, better preserving temporal information. As shown in Fig.~\ref{fig_1}(c), the TFM generates its output through the following operations:
\begin{equation}
  \label{X_m}
  X_m=concat\left( X_{l-1}^{1},X_{l-1}^{2} \right) .
\end{equation}
\begin{equation}
  \label{X_s}
  X_s=Conv_{1\times 1}\left( X_m \right) .
\end{equation}
\begin{equation}
  \label{T_l_}
  T_l=GELU\left( LN\left( X_m \right) \right) .
\end{equation}
Where, $X_{l-1}^{1},X_{l-1}^{2}\in R^{H\times W\times C_{in}}$ represent the input features, $X_m\in R^{H\times W\times 2C_{in}}$ denotes the features reduced through a pointwise convolution, and $T_l\in R^{H\times W\times C_{in}}$ stands for the temporally fused dual-temporal features. This approach ensures the retention of temporal information embedded in the dual-temporal features.
\subsection{Lightweight Feature Enhancement Module}\label{subsec:LFEM}
As model size increases, so does deployment complexity. To address this, we propose the LFEM, as shown in Fig.~\ref{fig_1}(d), which enhances feature extraction while maintaining computational and parameter efficiency. LFEM is based on the Mobile Inverted Bottleneck Convolution module~\cite{sandler2018mobilenetv2}. It starts with a $1\times 1$ convolution to make the feature space bigger, and then it moves on to a $3\times 3$ depth-wise convolution to make spatial interaction. It includes a Squeeze-and-Excitation (SE) module~\cite{hu2018squeeze} to adaptively reweight feature maps and concludes with a second $1\times 1$ convolution to project into the desired output channels. Mathematically, the LFEM at the $l$-th stage is expressed as follows:
\begin{equation}
  \label{F_e}
  F_{e}^{n}=Conv_{1\times 1}\left( F_{l-1}^{n} \right) .
\end{equation}
\begin{equation}
  \label{F_f}
  F_{f}^{n}=SE\left( DWConv_{3\times 3}\left( F_{l-1}^{n} \right) \right) .
\end{equation}
\begin{equation}
  \label{F_l}
  F_{l}^{n}=Conv_{1\times 1}\left( F_{f}^{n} \right) \oplus F_{l-1}^{n}.
\end{equation}
Where, $F_{l-1}^{n}\in R^{H\times W\times C_{in}}$ represents the input features, and $\mathrm{ }F_{e}^{n}\in R^{H\times W\times 6C_{in}}$ denotes the expanded features produced by the initial $1\times 1$ convolution. The output features are expressed as $F_{l}^{n}\in R^{H\times W\times C_{out}}$, where, $n=1,2$ corresponds to the features from the first and second temporal states, respectively. The operator $\oplus $ represents element-wise addition. The first two convolutional layers are followed by batch normalization and the SiLU activation function. When $C_{in}=C_{out}$, residual connections are applied to shorten the gradient propagation path, and Droppath~\cite{huang2016deep} is employed to stochastically skip network paths, introducing random depth to reduce complexity and prevent overfitting in the model.
\subsection{Multi-Scale Fusion Segmentation Head}\label{subsec:MSFSH}
To achieve multi-scale predictive outputs and address the challenge of vanishing gradients in backpropagation through shallow layers, we introduce the MSFSH. Similar to U2-Net~\cite{qin2020u2}, a $3\times 3$ convolution is applied at each decoder layer to generate prediction maps at multiple scales. After generating these prediction maps, they are resized to the original image dimensions using bilinear interpolation. Finally, a $1\times 1$ convolution is employed to fuse the multi-scale predictions, producing the final output.

\begin{figure*}[!t]
  \centering
  \includegraphics[width=6.5in]{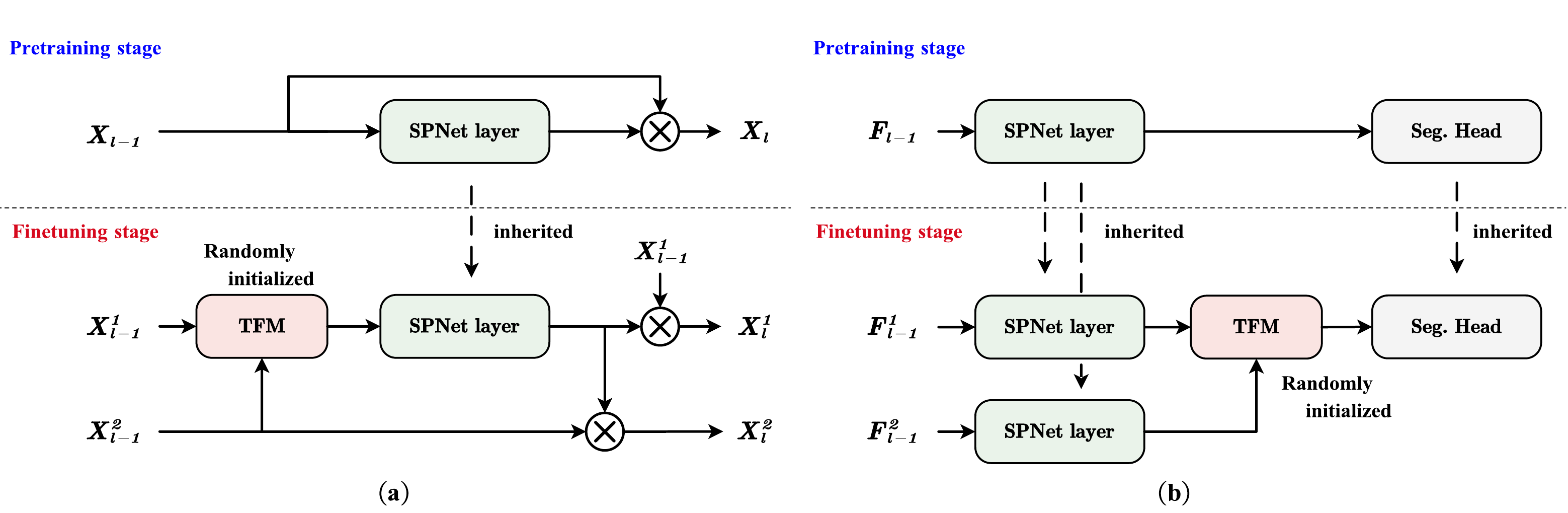}
  \caption{Overview of SCN. (a) represents the Shared Attention Fusion, while (b) represents the Siamese Feature Adaptation. During the fine-tuning stage, the weights of the SPNet layers are inherited from the pre-training stage, and the weights of the TFM are randomly initialized.}
  \label{fig_SCN}
  \end{figure*}
\subsection{Semantic Prior Network}
Viewing CD as a subset of the instances extraction task, its primary objective is to precisely identify and extract instances that have undergone alterations while disregarding instances that remain unchanged. To develop an effective model for this task, the network must first acquire prior knowledge in instance extraction. 

To effectively utilize prior knowledge from instance extraction, we design and pre-train SPNet on a single-temporal supervised task. SPNet integrates LFEM (Section~\ref{subsec:LFEM}) and Visual Attention Network Module (VANM)\cite{guo2023visual}, with MSFSH (Section~\ref{subsec:MSFSH}) generating the final prediction mask. As shown in Fig.~\ref{fig-BPnet}, SPNet follows a U-shaped structure\cite{ronneberger2015u}, consisting of a five-stage encoder, a five-stage decoder, VANM, and MSFSH. The network is constructed as follows:

Both the encoder and decoder consist of two sequential LFEMs. The first adjusts input feature dimensionality, while the second extracts deeper, high-level features. An initial feature extraction module with $3\times3$ convolution, BN, and SiLU activation is used before the first encoder block to expand feature dimensions. The encoder uses max pooling for downsampling, while the decoder uses bilinear interpolation for upsampling, thereby reducing both parameter count and computational complexity.

Position the VANM between the encoder and decoder to refine deeper features. For Stage 5, the decoder receives features processed by the VANM. For Stages 1 to 4, the upsampled feature map and the corresponding skip connection processed by VANM are added together to restore spatial details.

After each decoder stage, the feature maps are retained and processed through the MSFSH, with the final output generated using the sigmoid activation function.
\begin{figure*}[!t]
  \centering
  \includegraphics[width=7.0in]{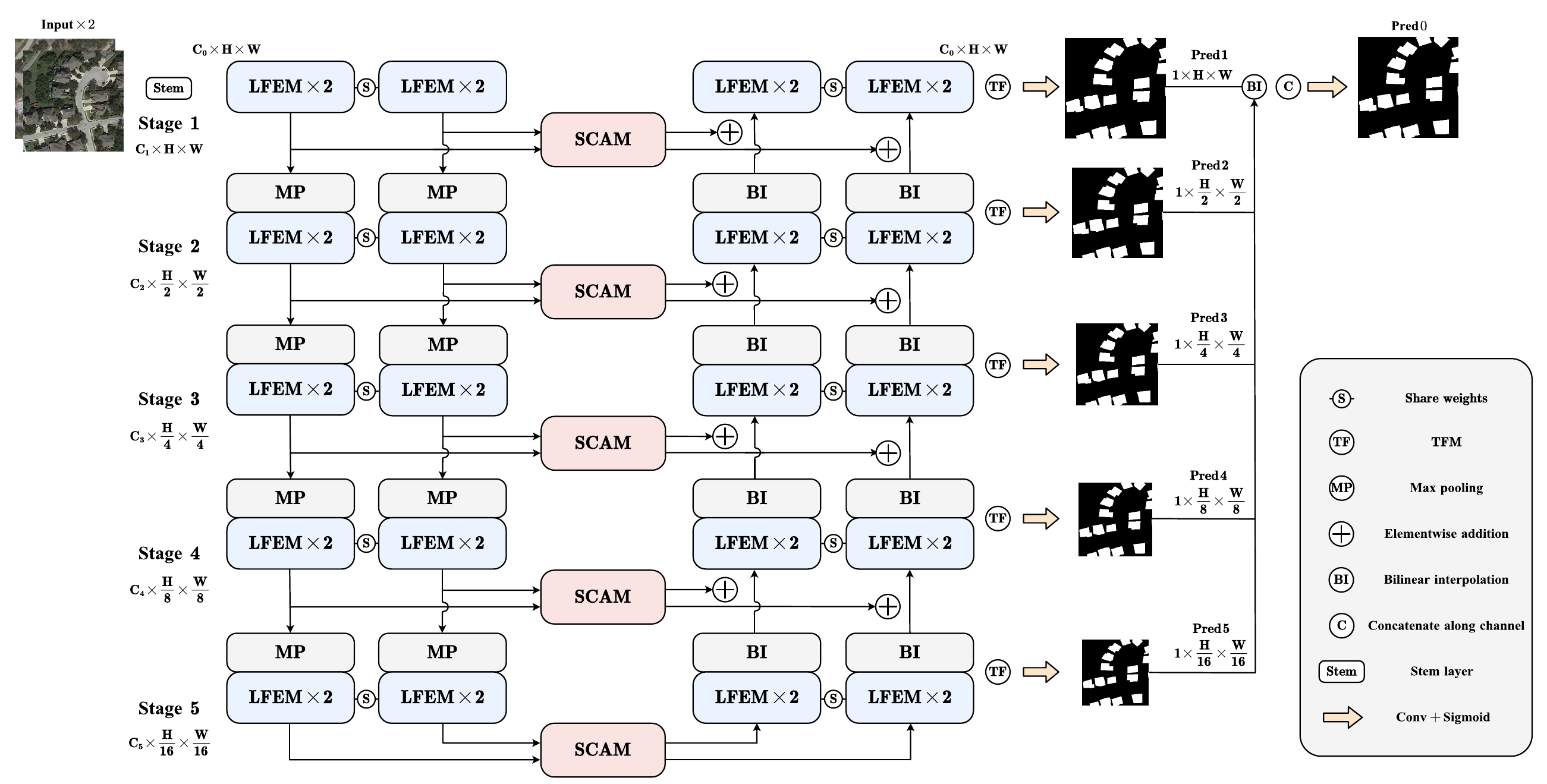}
  \caption{Overall Structure of SChanger. LFEM denotes the Lightweight Feature Enhancement Module (Section~\ref{subsec:LFEM}), while SCAM denotes the Spatial Consistency Attention Module (Section~\ref{subsec:SCAM}).}
  \label{fig-SChanger}
\end{figure*}
\subsection{Semantic Change Network}

  Semantic Change refers to the process of framing the CD task as a binary instance classification problem. After acquiring prior knowledge for instance extraction, the model focuses on determining whether the detected instances have undergone change. This approach simplifies the task, making it easier for the model to learn.

  To further enhance the model's ability to handle this task, we propose a fine-tuning strategy specifically designed for CD, called the Semantic Change Network (SCN). Through pretraining SPNet on a single-temporal supervised task, we observe that while SPNet exhibited strong generalization capabilities, it lacks the ability to effectively capture temporal changes. To improve its performance in CD tasks, we expand the model's capacity, adapting it to better handle the intricacies of detecting changes across time.

  We assume the capacity of SPNet to be fixed, consisting of $K$ layers $L_k$ for $k=1\cdots K$. Each layer contains hidden features $c_k\in R^{n_k}$, where $n_k$ represents the number of units in the $k$-th layer. Let $W_k$  denote the weight matrix (including bias terms) between layers $L_k$ and $L_{k-1}$, such that $c_k=f\left( W_kc_{k-1} \right) $, where $f\left( \cdot \right) $ is a nonlinear activation function, such as the SiLU. To construct the change-augmented representation module $F_c$, SCN introduces a new layer, $L_c$ (e.g., TFM), appended after the original network layers $L_k$, to enhance the model's capacity for CD. We consider $L_c$ as a CD adaptation layer, enabling new combinations of existing feature transformations without significantly modifying the pre-trained layers, thus adapting the model to the specific requirements of CD. The steps for SCN are as follows:

  Extending the Encoder and Decoder Architecture. To handle dual-temporal image inputs in CD tasks, we extend the encoder and decoder architecture using a shared-weight Siamese network. This setup allows the same encoder and decoder to process images from two different time points, ensuring both inputs utilize identical pre-trained weights. The Siamese network extracts consistent feature representations from both temporal instances. Specifically, for images from $t^1$ and $t^2$, the encoder and decoder generate feature representations $c_k^1=F_o\left( t^1 \right)$ and $c_k^2=F_o\left( t^2 \right)$, ensuring feature consistency and enhancing performance in CD tasks.

  Dual-Temporal Information Processing. To effectively capture the relationships between two distinct time points in change detection (CD) tasks, the shared-weight network must be capable of performing differential fusion. This is achieved through two primary mechanisms: (1) Shared Attention Fusion (SAF): As illustrated in Fig.~\ref{fig_SCN}(a), features from two separate time points, $X_1$ and $X_2$, are merged using a randomly initialized TFM. The fused features are subsequently processed through pretrained SPNet layers (such as VANM), which generate an attention map. This attention map is then applied to the feature maps from both time points, thereby enhancing its suitability for the CD task. Siamese Feature Adaptation (SFA): As shown in Fig.~\ref{fig_SCN}(b), in the extended architecture leveraging the Siamese network, features $F_1$ and $F_2$ are processed through SPNet layers, which retain pretrained weights. A randomly initialized TFM then reduces the feature dimensions by half to match the input required for the segmentation head. Together, these mechanisms form the CD adaptation layer, $L_c$, enhancing the model's ability to adapt to CD tasks. This process is expressed mathematically as $c_c = F_c\left( F_o\left( t^1 \right), F_o\left( t^2 \right) \right)$.

  Through these refinements, the model has successfully integrated dual-temporal information processing capabilities while maintaining its structural stability. Despite the model's expansion and the introduction of new modules, the overall parameter count and computational complexity remain relatively modest, keeping the total fine-tuning cost within acceptable bounds. The SChanger model is referred to as the SPNet augmented by SCN. As shown in Fig.~\ref{fig-SChanger}, SChanger is designed to capture dual-temporal difference representations while minimizing computational and parameter overhead. By leveraging SCN, the pre-trained weights are seamlessly incorporated into the CD architecture with only a minimal increase in parameters. This enables the network to learn more robust instance features, ultimately improving performance.

  In this paper, we present two Variants of SChanger by configuring different numbers of filters: the standard SCchanger-base and the comparatively smaller SChanger-small. Detailed configuration profiles are provided in Table~\ref{tab:SChanger_summary}.

  \begin{table}[!t]
    \caption{Network Architecture for SChanger Variants. $\Delta$Parameters indicates the increase in parameter count resulting from the application of the SCN strategy.\label{tab:SChanger_summary}}
    \centering
    \resizebox{\columnwidth}{!}{%
      \begin{tabular}{c|l|l|l}  
        \toprule
        \textbf{Stage} & \textbf{downsample ratio} & \textbf{SChanger-small} & \textbf{SChanger-base} \\
        \midrule
        stem & 1 & $C_0=8$ & $C_0=24$ \\
        1    & 1 & $C_1=16$ & $C_1=32$ \\
        2    & 2 & $C_2=32$ & $C_2=48$ \\
        3    & 4 & $C_3=40$ & $C_3=64$ \\
        4    & 8 & $C_4=48$ & $C_4=104$ \\
        5    & 16 & $C_5=48$ & $C_5=120$ \\
        \midrule
        \textbf{Flops (G)}        &  & $6.242$ & $18.275$ \\
        \textbf{Parameters (M)}  &  & $0.607$ & $2.370$ \\
        \rowcolor{mygray}  
        \textbf{$\Delta$Parameters (M)} &  & $0.026$ (4.2\%) & $0.105$ (4.4\%) \\
        \bottomrule
      \end{tabular}%
    }
\end{table}

\subsection{Training Details}
To enhance SChanger's capability to process multi-scale information and reduce back-propagation distance, we incorporate Deep Supervision~\cite{qin2020u2} to improve the model's prediction accuracy. Deep Supervision calculates the loss function at each layer of the network, generating multi-scale mask outputs. Since the datasets used during both pretraining and fine-tuning stages contain a single label, the same loss function is applied consistently across stages. The loss function for each layer, as well as the total loss function, is expressed as:
\begin{equation}
  \label{l_i}
  l_i=\text{Bce}\left( y,\hat{y} \right) +\text{Dice}\left( y,\hat{y} \right).
\end{equation}
\begin{equation}
  \label{loss}
  Loss=\sum_{i=0}^5{\lambda _i\times l_i}.
\end{equation}
Where $\text{Bce}\left( \cdot\right)$ and $\text{Dice}\left( \cdot \right)$ represent binary cross-entropy and Dice loss, respectively. The term $\lambda_i$ denotes the loss function coefficients for different layers, which we set to 1 in this study. Additionally, to improve model stability, we adopt the Exponential Moving Average~\cite{he2020momentum} model with synchronized updates across layers.
Formally, let the parameters of the model be denoted as \( \theta_q \), and the updated parameters stored separately as \( \theta_k \). The update rule for \( \theta_k \) is given by:
\begin{equation}
  \label{EMA}
  \theta_\mathrm{k}\leftarrow m\theta_\mathrm{k}+(1-m)\theta_\mathrm{q}.
\end{equation}
Here, \( m \in [0, 1) \) is the momentum coefficient, set to a commonly used value (e.g., \( m = 0.9998 \)).

For models with fewer parameters, fully fine-tuning all parameters has been demonstrated to yield better results~\cite{wang2017growing}. Building on this, during the CD phase, we fully update the pre-trained weights to maximize model's effectiveness and further enhance its performance on the new task.
\section{Experiments}
\subsection{Datasets}
\begin{table*}[!t]
  \caption{Information of the Seven Benchmark Datasets Used for Experiments. \label{tab:benchmark_datasets}}
  \centering
    \begin{tabular}{l|l|c|c|c|l}  
      \toprule
      \textbf{Dataset} & \textbf{Study Site} & \textbf{Number of Samples} & \textbf{Size} & \textbf{Resolution} & \textbf{Evaluation Task} \\
      \midrule
      LEVIR-CD~\cite{chen2020spatial}        & Texas                   & 637   & 1024 $\times$ 1024  & 0.5m  & object change detection \\
      LEVIR-CD+~\cite{chen2020spatial}   & Texas                  & 985      & 1,024 $\times$ 1,024 & 0.5m  & object change detection \\
      S2looking~\cite{shen2021s2looking}   & Rural regions worldwide  & 5,000    & 1,024 $\times$ 1,024 & 0.5-0.8m & object change detection \\
      CDD~\cite{lebedev2018change}         & -       & 16000      & 256 $\times$ 256    & 0.03-1m & binary change detection  \\
      WHU-CD~\cite{ji2018fully}      & Christchurch, New Zealand    & 1        & 32207 $\times$ 15354 & 0.3m  & object change detection \\
      SYSU-CD~\cite{shi2021deeply}      & Hong Kong  & 20000    & 256 $\times$ 256  & 0.5m & binary change detection \\
      \midrule
      IAILD & Austin, Chicago, Kitsap, Tyrol, Vienna  & 360    & 5000 $\times$ 5000  & 0.3m  & building extraction \\
      \bottomrule
    \end{tabular}
\end{table*}
The basic information of the seven datasets used in the experiments is summarized in Table~\ref{tab:benchmark_datasets}. The following provides the details of each dataset:

The LEVIR-CD dataset~\cite{chen2020spatial}, widely used for CD, contains 637 pairs of very high-resolution image patches (0.5 meters per pixel) from Google Earth, each measuring $1024\times 1024$ pixels. It features 31,333 annotated instances of building changes, with imagery captured between 2002 and 2018. The dataset primarily focuses on transitions from grassland, soil, or construction sites to developed buildings. For evaluation, the test set images are cropped to $256\times 256$ pixels.

LEVIR-CD+ dataset includes 985 pairs of images collected between 2002 and 2020, encompassing approximately 80,000 building instances. For consistency, we apply the same cropping methodology as in LEVIR-CD in our experiments.

The S2Looking dataset~\cite{shen2021s2looking} includes 5,000 image pairs, each with dimensions of $1024 \times 1024$ pixels, and is divided into training, validation, and test sets with a 7:1:2 ratio. It features more than 65,920 annotated change instances, derived from side-looking satellite images of rural areas globally, with spatial resolutions between 0.5 and 0.8 meters per pixel. The dataset presents challenges such as large viewing angles, significant illumination variations, and the complexity of rural imagery. Like the LEVIR-CD dataset, the test images are cropped to $256 \times 256$ pixels for evaluation.

The CDD dataset~\cite{lebedev2018change} includes 11 pairs of Google Earth images, where seasonal variations impact mutable objects such as buildings, roads, and vehicles. This dataset poses additional challenges due to seasonal and lighting differences. The dataset is split into patches of $256\times 256$ pixels, with 10,000 patches for training, 3,000 for validation, and 3,000 for testing.

The WHU-CD dataset~\cite{ji2018fully} is a large-scale resource specifically created for building change detection. It consists of high-resolution image pairs, each measuring $32207\times 15354$ pixels, taken in Christchurch, New Zealand, beginning in 2011. With a spatial resolution of 0.2 meters per pixel, the dataset captures significant post-earthquake changes, particularly related to building reconstruction. To enable fair comparisons with other algorithms, the images were cropped into non-overlapping blocks of $256\times 256$ pixels.

The SYSU-CD dataset~\cite{shi2021deeply}, based in Hong Kong, contains 20,000 images with a resolution of $256 \times 256$ pixels and a spatial resolution of 0.5 meters per pixel. The dataset captures a diverse array of complex change scenarios, including road expansions, the development of new urban buildings, vegetation changes, suburban growth, and groundwork before construction. 

The Inria Aerial Image Labeling Dataset (IAILD)~\cite{maggiori2017can} comprises 360 images, each with a resolution of $5000\times 5000$ pixels, collected from five cities: Austin, Chicago, Kitsap, Tyrol, and Vienna. In this study, we utilize the entire training set for pretraining SPNet on the building extraction task, as the test set lacks labeled data.

\subsection{Implementation Detail}
To assess the performance of the proposed model, all experiments are conducted in a PyTorch 2.6~\cite{ansel2024pytorch} environment (CUDA 11.1), utilizing an NVIDIA GeForce RTX 4090 GPU with 24GB of memory. TorchInductor~\cite{tillet2019triton} is employed as the deep learning compiler to generate optimized code, accelerating both training and inference speeds. We employ the AdamW optimizer~\cite{loshchilov2017fixing} with an initial learning rate of 0.0005, a cosine decay schedule, and a weight decay of 0.0002 to prevent overfitting. Data augmentation techniques include random cropping to $256\times 256$ pixels, horizontal/vertical flips (50\% probability), random rotations, translations, and scaling (30\% probability). Additional augmentations such as contrast, gamma corrections, emboss effects, Gaussian noise, adjustments to hue, saturation, brightness, and motion blur are applied with a 50\% probability. For CD datasets, dual-temporal images are randomly swapped with a 50\% probability. The model is trained across multiple datasets for rigorous evaluation performance.

For IAILD, we start with 400 warmup epochs and a batch size of 16, followed by 4000 training epochs. For LEVIR-CD, the configuration includes 600 warmup epochs and a total of 12,000 epochs. LEVIR-CD+ requires 450 warmup epochs and 9000 total epochs. S2Looking involves 150 warmup epochs and 3000 total epochs. CDD uses 50 warmup epochs and 500 total epochs. SYSU-CD utilizes 10 warmup epochs and 450 total epochs. WHU-CD consists of 20 warmup epochs and 1000 total epochs. For all CD tasks, the models are initialized with pre-trained weights from IAILD, and a batch size of 8 is employed.
\subsection{Evaluation Metrics}
For the CD task, we take the F1 score of the change class as the primary evaluation metric. The F1 score, along with precision and recall, are calculated as follows to evaluate the model's performance:
\begin{equation}
  \label{F_1}
  \text{F1}=\frac{2\times \text{Precision}\times \text{Recall}}{\text{Precision}+\text{Recall}}.
\end{equation}
\begin{equation}
  \label{Precision}
  \text{Precision}=\frac{\text{TP}}{\text{TP}+\text{FP}}.
\end{equation}
\begin{equation}
  \label{recall}
  \text{Recall}=\frac{\text{TP}}{\text{TP}+\text{FN}}.
\end{equation}
Where TP denotes true positives, TN refers to true negatives, FP indicates false positives, and FN represents false negatives.
\subsection{Compared Methods}
To evaluate the performance of the proposed SChanger model, we compare it against several leading CD methods. Current CD methods in deep learning fall into three major categories: models based on CNNs, models based on convolutional attention mechanisms, and models based on transformer architectures. For the CNN-based methods, we select several classic and SOTA algorithms for comparison, including FC-EF~\cite{daudt2018fully}, FC-Siam-Conc~\cite{daudt2018fully}, FC-Siam-Diff~\cite{daudt2018fully}, UNet++MSOF~\cite{peng2019end}, ChangeStar2~(R-50)~\cite{zheng2024single}, and ChangerEx~\cite{fang2023changer}. These methods rely on CNNs to extract spatial features for accurate CD. For the attention-based convolutional methods, we include models like DTCDSCN~\cite{liu2020building}, IFN~\cite{zhang2020deeply}, SGSLN/512~\cite{zhao2023exchanging},HANet~\cite{han2023hanet}, CGNet~\cite{han2023change}, C2FNet~\cite{han2024c2f}, SRCNet~\cite{chen2024src}, STANet~\cite{chen2020spatial}, CACG-Net~\cite{liu2024candidate}, and Intelligent-BCD~\cite{zhang2022intelligent}. These models integrate attention mechanisms to enhance accuracy and robustness by focusing on the most relevant features. Among transformer-based approaches, we select ChangeFormer~\cite{bandara2022transformer}, BiT~\cite{chen2021remote}, MutSimNet~\cite{liu2024mutsimnet}, and TransUNetCD~\cite{li2022transunetcd}. These methods leverage the self-attention mechanism of transformers to capture long-range dependencies in images, which significantly enhances their performance in CD tasks.

\begin{table}[!t]
  \caption{Comparison of the Results With Other SOTA Methods on LEVIR-CD. Color convention:~First~(red), Second~(blue) and Third~(bold).\label{tab:result-LEVIR-CD}}
  \centering
  \begin{tabular}{l|c|ccc}
    \toprule
    \textbf{Methods} & \textbf{Year} & \textbf{P. (\%)} & \textbf{R. (\%)} & \textbf{F1 (\%)} \\
    \midrule
    FC-EF~\cite{daudt2018fully} & 2018 & 86.91 & 80.17 & 83.40 \\
    FC-Siam-Conc~\cite{daudt2018fully} & 2018 & 91.99 & 76.77 & 83.69 \\
    FC-Siam-Diff~\cite{daudt2018fully} & 2018 & 89.53 & 83.31 & 86.31 \\
    DTCDSCN~\cite{liu2020building} & 2020 & 88.53 & 86.83 & 87.67 \\
    ChangeFormer~\cite{bandara2022transformer} & 2020 & 92.59 & 89.68 & 91.11 \\
    IFN~\cite{zhang2020deeply} & 2020 & \textcolor{red}{\textbf{94.02}} & 82.93 & 88.13 \\
    BiT-18~\cite{chen2021remote} & 2021 & 89.24 & 89.37 & 89.31 \\
    HANet~\cite{han2023hanet}& 2023 & 91.21 & 89.36 & 90.28 \\
    ChangerEx~\cite{fang2023changer} & 2023 & 92.97 & 90.61 & 91.77 \\
    SGSLN/512~\cite{zhao2023exchanging} & 2023 & 93.07 & 91.61 & 92.33 \\
    LightCDNet~\cite{xing2023lightcdnet}& 2023 & 92.43 & 90.45 & 91.43 \\
    CGNet~\cite{han2023change} & 2023 & 93.15 & 90.90 & 92.01 \\
    C2FNet~\cite{han2024c2f} & 2024 & \textcolor{softblue}{\textbf{93.69}} & 90.04 & 91.83 \\
    SRCNet~\cite{chen2024src} & 2024 & 92.63 & \textbf{91.72} & 92.24 \\
    CACG-Net~\cite{liu2024candidate} & 2024 & 92.16 & 92.41 & 92.29  \\
    MutSimNet~\cite{liu2024mutsimnet} & 2024 & 90.46 & 93.59 & 92.00  \\
    MTP~\cite{wang2024mtp}& 2024 & - & - & \textcolor{softblue}{\textbf{92.67}} \\
    SChanger-small & - & 92.93 & \textcolor{softblue}{\textbf{91.96}} & \textbf{92.45} \\
    SChanger-base & - & \textbf{93.18} & \textcolor{red}{\textbf{92.61}} & \textcolor{red}{\textbf{92.87}} \\
    \bottomrule
  \end{tabular}
\end{table}
\subsection{Main results}
\begin{figure*}[!t]
  \centering
  \includegraphics[width=6.5in]{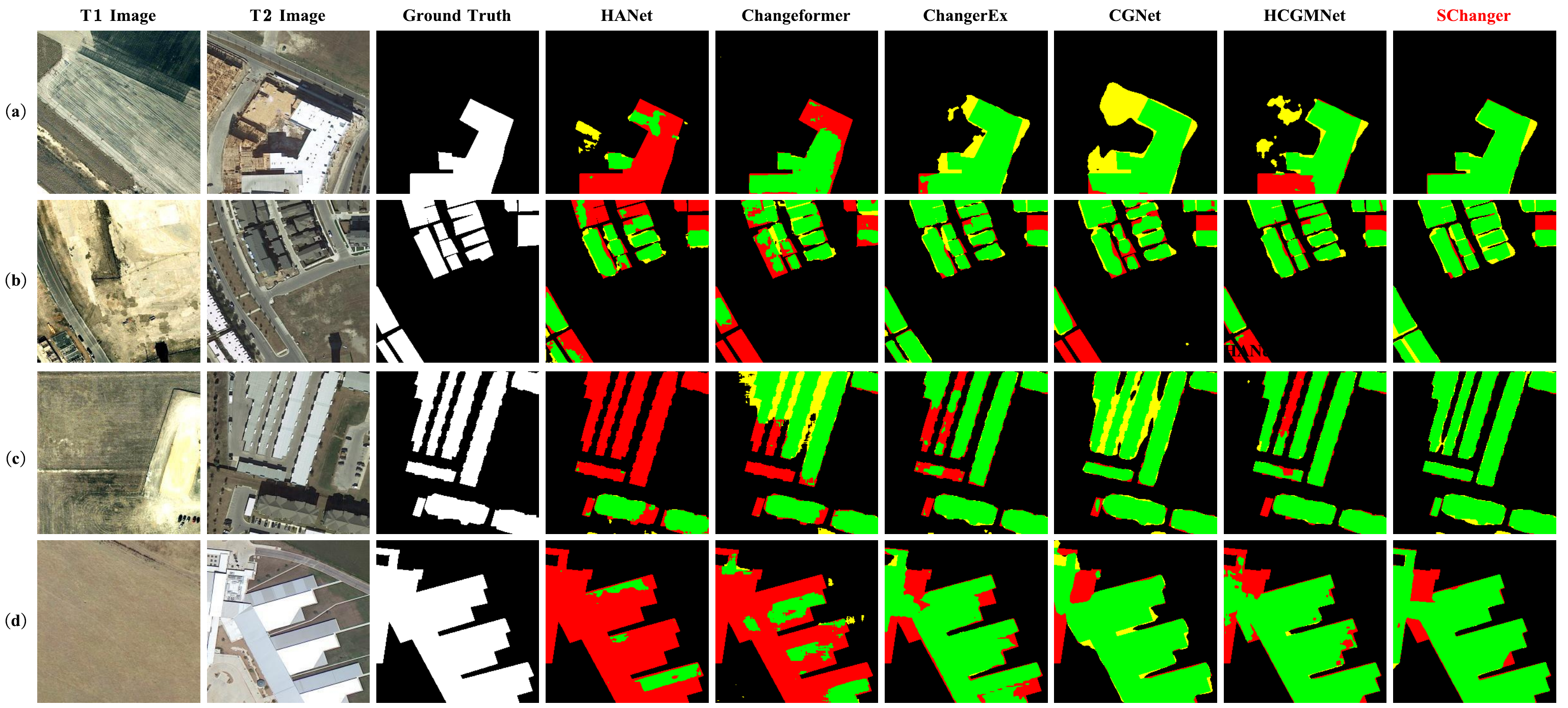}
  \caption{Qualitative evaluation results of SOTA comparison methods on the LEVIR-CD dataset. TP (green), TN (black), FP (yellow), and FN (red).}
  \label{fig-othermodellevir}
\end{figure*}

The experimental results on the LEVIR-CD dataset, as shown in Table~\ref{tab:result-LEVIR-CD}, highlight the superior performance of the proposed SChanger model. SChanger-small achieves an F1 score of 92.45\%, and SChanger-base reaches 92.87\%. These results represent the SOTA performance on this dataset. Compared to the previous best-performing model, MTP, SChanger improves the F1 score by 0.20\%. Additionally, it outperforms the previous best lightweight model, SGSLN/512, improving precision, recall, and F1 score by 0.11\%, 1.00\%, and 0.54\%, respectively. Figs.~\ref{fig-othermodellevir} present the detection results for some examples from the LEVIR-CD dataset. It is evident that the outcomes produced by our proposed model align more closely with the corresponding ground truth compared to the results from other models.

Table~\ref{tab:result-LEVIR-CD2} shows the comparison of accuracy results 
on the LEVIR-CD+ dataset. SChanger-small records an F1 score of 86.20\%, with SChanger-base raising it to 86.43\%. These results outperform the CNN-based Intelligent-BCD by 0.14\% and the Transformer-based BiT by 3.64\%. Despite LEVIR-CD+ being more challenging than LEVIR-CD, SChanger-base's performance only drops by 6.44\% and SChanger-small's by 6.25\%, outperforming models like CGNet (8.33\%), BiT (7.47\%), and DTCDSCN (10.07\%). This highlights SChanger's robustness and adaptability in handling complex changes.
\begin{table}[!t]
  \caption{Comparison of the Results With Other SOTA Methods on LEVIR-CD+. Color convention:~First~(red), Second~(blue) and Third~(bold).\label{tab:result-LEVIR-CD2}}
  \centering
  \begin{tabular}{l|c|ccc}
    \toprule
    \textbf{Methods} & \textbf{Year} & \textbf{P. (\%)} & \textbf{R. (\%)} & \textbf{F1 (\%)} \\
    \midrule
    FC-EF~\cite{daudt2018fully} & 2018 & 61.30 & 72.61 & 66.48 \\
    FC-Siam-Conc~\cite{daudt2018fully} & 2018 & 66.24 & 81.22 & 72.97 \\
    FC-Siam-Diff~\cite{daudt2018fully} & 2018 & 74.97 & 72.04 & 73.48 \\
    UNet++MSOF~\cite{peng2019end} & 2019 & \textcolor{softblue}{\textbf{85.90}} & 67.10 & 75.34 \\
    DTCDSCN~\cite{liu2020building} & 2020 & 80.36 & 75.03 & 77.60 \\
    STANet~\cite{chen2020spatial} & 2020 & 74.62 & 84.54 & 79.27 \\
    BiT-18~\cite{chen2021remote} & 2021 & 82.74 & 82.85 & 82.79 \\
    Intelligent-BCD~\cite{zhang2022intelligent} & 2022 & \textcolor{red}{\textbf{93.80}} & 79.90 & \textcolor{softblue}{\textbf{86.29}} \\
    CGNet~\cite{han2023change} & 2023 & 81.46 & \textbf{86.02} & 83.68 \\
    SChanger-small & - & 84.53 & \textcolor{red}{\textbf{87.93}} & \textbf{86.20} \\
    SChanger-base & - & \textbf{85.28} & \textcolor{softblue}{\textbf{87.61}} & \textcolor{red}{\textbf{86.43}} \\
    \bottomrule
  \end{tabular}
\end{table}

\begin{figure*}[!t]
  \centering
  \includegraphics[width=6in]{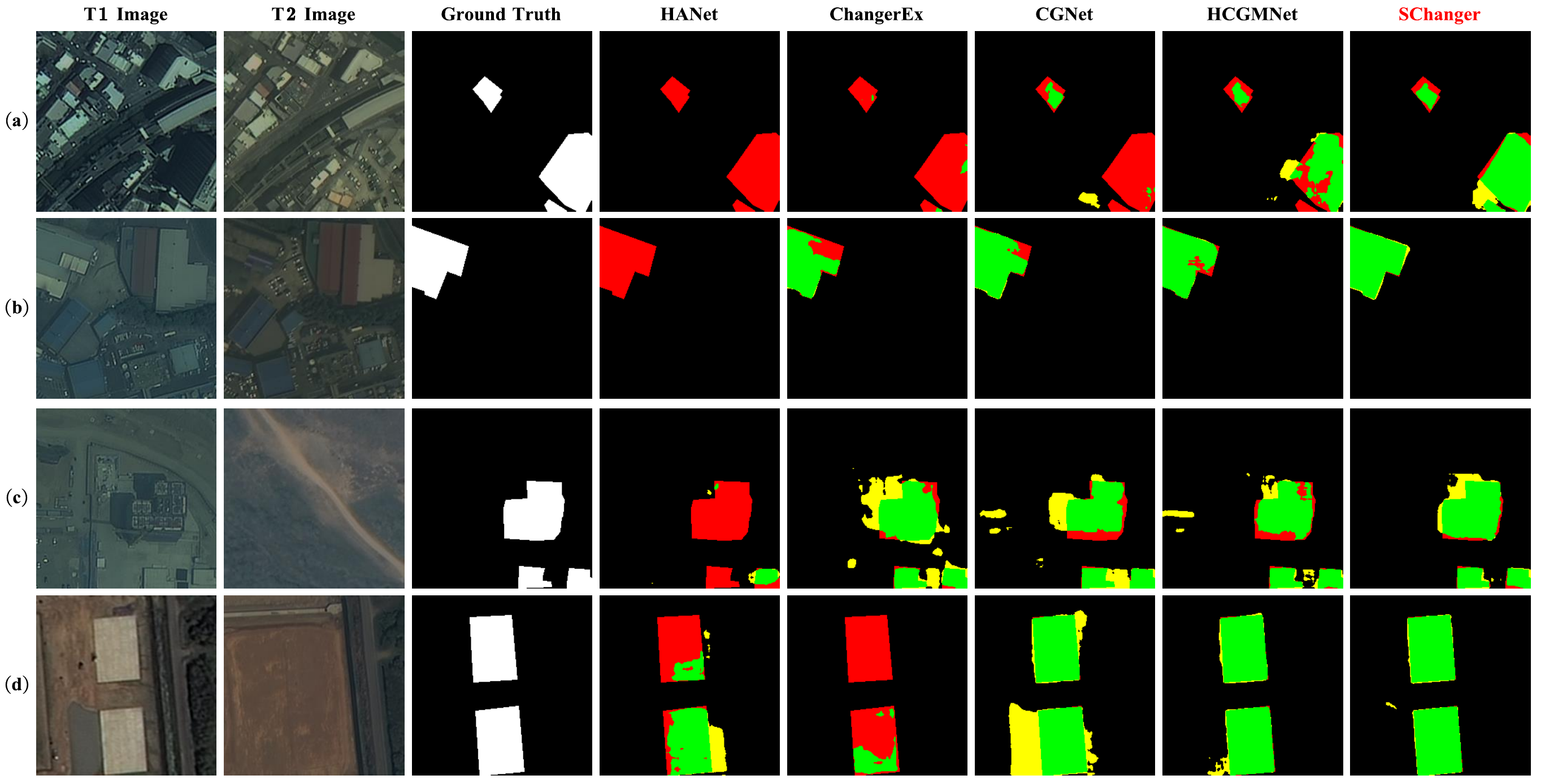}
  \caption{Qualitative evaluation results of SOTA comparison methods on the S2looking dataset. TP (green), TN (black), FP (yellow), and FN (red).}
  \label{fig-othermodels2looking}
\end{figure*}

As shown in Table~\ref{tab:result-s2looking}, our method performs strongly on the S2Looking dataset. SChanger-base achieves an F1 score of 68.95\% and SChanger-small scores 68.20\%, surpassing the previous SOTA method, ChangeStar2, by 1.15\% and 0.40\%, respectively. Due to significant visual differences in rural bitemporal images, the S2Looking dataset presents challenges. SChanger's prior knowledge of building extraction from IAILD and spatial consistency inductive bias enables more accurate, efficient detection change, resulting in improved performance on these complex scenes. Figs.~\ref{fig-othermodels2looking} show detection results from the s2looking dataset, where our model's outcomes align more closely with the ground truth than those of other models.

\begin{table}[!t]
  \caption{Comparison of the Results With Other SOTA Methods on S2looking. Color convention:~First~(red), Second~(blue) and Third~(bold).\label{tab:result-s2looking}}
  \centering
  \begin{tabular}{l|c|ccc}
    \toprule
    \textbf{Method} & \textbf{Year} & \textbf{P. (\%)} & \textbf{R. (\%)} & \textbf{F1 (\%)} \\
    \midrule
    FC-EF~\cite{daudt2018fully} & 2018 & \textcolor{softblue}{\textbf{81.36}} & 8.95  & 7.65  \\
    FC-Siam-Conc~\cite{daudt2018fully} & 2018 & 68.27 & 18.52 & 13.54 \\
    FC-Siam-Diff~\cite{daudt2018fully} & 2018 & \textcolor{red}{\textbf{83.29}} & 15.76 & 13.19 \\
    ChangeFormer~\cite{bandara2022transformer} & 2020 & 72.82 & 56.13 & 63.39 \\
    DTCDSCN~\cite{liu2020building} & 2020 & 68.58 & 49.16 & 57.27 \\
    BiT-18~\cite{chen2021remote} & 2021 & 72.64 & 53.85 & 61.85 \\
    HANet~\cite{han2023hanet}& 2023 & 61.38 & 55.94 & 58.54 \\
    ChangerEx~\cite{fang2023changer} & 2023 & 73.59 & 60.15 & 66.20 \\
    CGNet~\cite{han2023change} & 2023 & 70.18 & 59.38 & 64.33 \\
    C2FNet~\cite{han2024c2f} & 2024 & \textbf{74.84} & 54.14 & 62.83 \\
    ChangeStar2~\cite{zheng2024single} & 2024 & 69.20 & \textcolor{softblue}{\textbf{66.50}} & \textbf{67.80} \\
    SChanger-small & - & 70.63 & \textbf{65.93} & \textcolor{softblue}{\textbf{68.20}} \\
    SChanger-base & - & 70.94 & \textcolor{red}{\textbf{67.09}} & \textcolor{red}{\textbf{68.95}} \\
    \bottomrule
  \end{tabular}
\end{table}

Table~\ref{tab:cdd} highlights the performance of our method on the CDD dataset. SChanger-small achieves an F1 score of 95.75\%, while SChanger-base reaches 97.62\%, reflecting a 0.12\% improvement over the second-best method, ChangeStar2. These results demonstrate the model's robustness in handling scenarios involving seasonal changes.

\begin{table}[!t]
  \caption{Comparison of the Results With Other SOTA Methods on CDD. Color convention:~First~(red), Second~(blue) and Third~(bold).\label{tab:cdd}}
  \centering
  \begin{tabular}{l|c|ccc}
    \toprule
    \textbf{Methods} & \textbf{Year} & \textbf{P. (\%)} & \textbf{R. (\%)} & \textbf{F1 (\%)} \\
    \midrule
    FC-EF~\cite{daudt2018fully} & 2018 & 60.90 & 58.30 & 59.57 \\
    FC-Siam-Conc~\cite{daudt2018fully} & 2018 & 70.90 & 60.30 & 65.17 \\
    FC-Siam-Diff~\cite{daudt2018fully} & 2018 & 76.20 & 57.30 & 65.41 \\
    UNet++MSOF~\cite{peng2019end} & 2019 & 86.68 & 76.53 & 81.29 \\
    IFN~\cite{zhang2020deeply} & 2020 & 90.56 & 70.18 & 79.08 \\
    HANet~\cite{han2023hanet}& 2023 & 92.89 & 85.87 & 89.23 \\
    CGNet~\cite{han2023change} & 2023 & 93.67 & 95.82 & 94.73 \\
    C2FNet~\cite{han2024c2f} & 2024 & 95.46 &\textbf{96.41} & \textbf{95.93} \\
    ChangeStar2~\cite{zheng2024single} & 2024 &  \textcolor{red}{\textbf{98.00}} & \textcolor{softblue}{\textbf{97.00}} & \textcolor{softblue}{\textbf{97.50}} \\
    SChanger-small & - & \textbf{95.48} & 96.02 & 95.75 \\
    SChanger-base & - & \textcolor{softblue}{\textbf{97.95}} & \textcolor{red}{\textbf{97.28}} & \textcolor{red}{\textbf{97.62}} \\
    \bottomrule
  \end{tabular}
\end{table}

Table~\ref{tab:sysu} shows the results on the SYSU dataset, which includes multiple categories. The comparison reveals that SChanger-small achieves the highest F1 score of 84.58\%, while SChanger-base follows with a score of 84.17\%, surpassing the previous best method, CACG-Net, by 1.23\%. It is noteworthy that SChanger-small outperforms SChanger-base, likely due to the simpler nature of the dataset. The smaller model is less prone to overfitting, allowing it to generalize better, while the larger model, with its increased complexity, may be more susceptible to overfitting.

\begin{table}[!t]
  \caption{Comparison of the Results With Other SOTA Methods on SYSU-CD. Color convention:~First~(red), Second~(blue) and Third~(bold).}\label{tab:sysu}
  \centering
  \begin{tabular}{l|c|ccc}
    \toprule
    \textbf{Methods} & \textbf{Year} & \textbf{P. (\%)} & \textbf{R. (\%)} & \textbf{F1 (\%)} \\
    \midrule
    FC-EF~\cite{daudt2018fully} & 2018 & 76.47 & 75.17 & 75.81 \\
    FC-Siam-Conc~\cite{daudt2018fully} & 2018 & 73.67 & 76.75 &  75.21 \\
    FC-Siam-Diff~\cite{daudt2018fully} & 2018 & 76.28 & 75.30 & 75.79 \\
    IFN~\cite{zhang2020deeply} & 2020 &  79.59 & 75.58 &  77.53 \\
    STANet~\cite{chen2020spatial} & 2020 &  70.76 & \textcolor{red}{\textbf{85.33}} & 77.36 \\
    HANet~\cite{han2023hanet}& 2023 & 78.71 & 76.14 & 77.41 \\
    CGNet~\cite{han2023change} &2023& \textcolor{red}{\textbf{86.37}} &74.37 &79.92\\
    SGSLN/512~\cite{zhao2023exchanging} & 2023 &  84.76 &  81.45 & 83.07 \\
    C2FNet~\cite{han2024c2f} & 2024 & 75.44 &80.67 & 77.97	\\
    MutSimNet~\cite{liu2024mutsimnet} & 2024 &  82.91 &  81.73 & 82.34 \\
    CACG-Net~\cite{liu2024candidate} & 2024 & 82.42 & \textcolor{softblue}{\textbf{84.30}} & \textbf{83.35}  \\
    SChanger-small & - & \textbf{85.05} & \textbf{84.11} & \textcolor{red}{\textbf{84.58}} \\
    SChanger-base & - & \textcolor{softblue}{\textbf{86.35}} & 82.09 & \textcolor{softblue}{\textbf{84.17}} \\
    \bottomrule
  \end{tabular}
\end{table}

 Table~\ref{tab:whu-cd} highlights the performance of our method on the WHU-CD dataset, which is widely used for change detection tasks. SChanger-small achieves an F1 score of 93.15\%, demonstrating its effectiveness in identifying and detecting changes across different categories in the dataset. Meanwhile, SChanger-base performs slightly better, with an F1 score of 93.20\%. Despite the minimal difference between the two models, the results showcase the robustness of our approach in handling complex tasks.

\begin{table}[!t]
  \caption{Comparison of the Results With Other SOTA Methods on WHU-CD. Color convention:~First~(red), Second~(blue) and Third~(bold).\label{tab:whu-cd}}
  \centering
  \begin{tabular}{l|c|ccc}
    \toprule
    \textbf{Methods} & \textbf{Year} & \textbf{P (\%)} & \textbf{R (\%)} & \textbf{F1 (\%)} \\
    \midrule
    FC-EF~\cite{daudt2018fully} & 2018 & 83.50 &  86.33 & 84.89 \\
    FC-Siam-Conc~\cite{daudt2018fully} & 2018 & 84.02 &  87.72 &  85.83 \\
    FC-Siam-Diff~\cite{daudt2018fully} & 2018 & 90.86 & 84.69 &  87.67 \\
    DTCDSCN~\cite{liu2020building} & 2020 & 63.92 &  82.30 & 71.95 \\
    IFN~\cite{zhang2020deeply} & 2020 &   91.44 &  89.75 &  90.59 \\
    BIT-18~\cite{chen2021remote}&2021& 90.30&90.36& 90.33\\
    HANet~\cite{han2023hanet}& 2023 & 88.30 & 88.01  & 88.16  \\
    CGNet~\cite{han2023change} &2023&  \textbf{94.47} & \textbf{90.79} &\textbf{92.59}\\
    SChanger-small & - & \textcolor{red}{\textbf{95.36}} & \textcolor{softblue}{\textbf{91.04}} & \textcolor{softblue}{\textbf{93.15}} \\
    SChanger-base & - & \textcolor{softblue}{\textbf{94.62}} & \textcolor{red}{\textbf{91.83}} & \textcolor{red}{\textbf{93.20}} \\
    \bottomrule
  \end{tabular}
\end{table}

\subsection{Few-shot Learning}

To evaluate SCN's impact on the generalization capability of SChanger, we conduct a few-shot learning experiment on the LEVIR-CD dataset, reducing the training samples from 30\% to 5\% and assessing the model's performance on the test set. 
Leveraging SCN, SChanger effectively acquires prior knowledge. As presented in Table~\ref{tab:fewshot_performance}, with 5\%, 10\%, 20\%, 30\%, and 100\% of the training samples, SChanger achieves F1 scores of 91.26\%, 91.57\%, 92.04\%, 92.47\%, and 92.87\%, respectively. These results underscore the critical role of SCN in enhancing SChanger's performance in few-shot learning, with significant F1 score improvements compared to random initialization, by 4.50\%, 2.03\%, 1.30\%, 1.22\%, and 0.33\%, further validating the effectiveness and importance of SCN.

\begin{figure}[!t]
  \centering
  \hspace{-0.1in}  
  \includegraphics[width=3.3in]{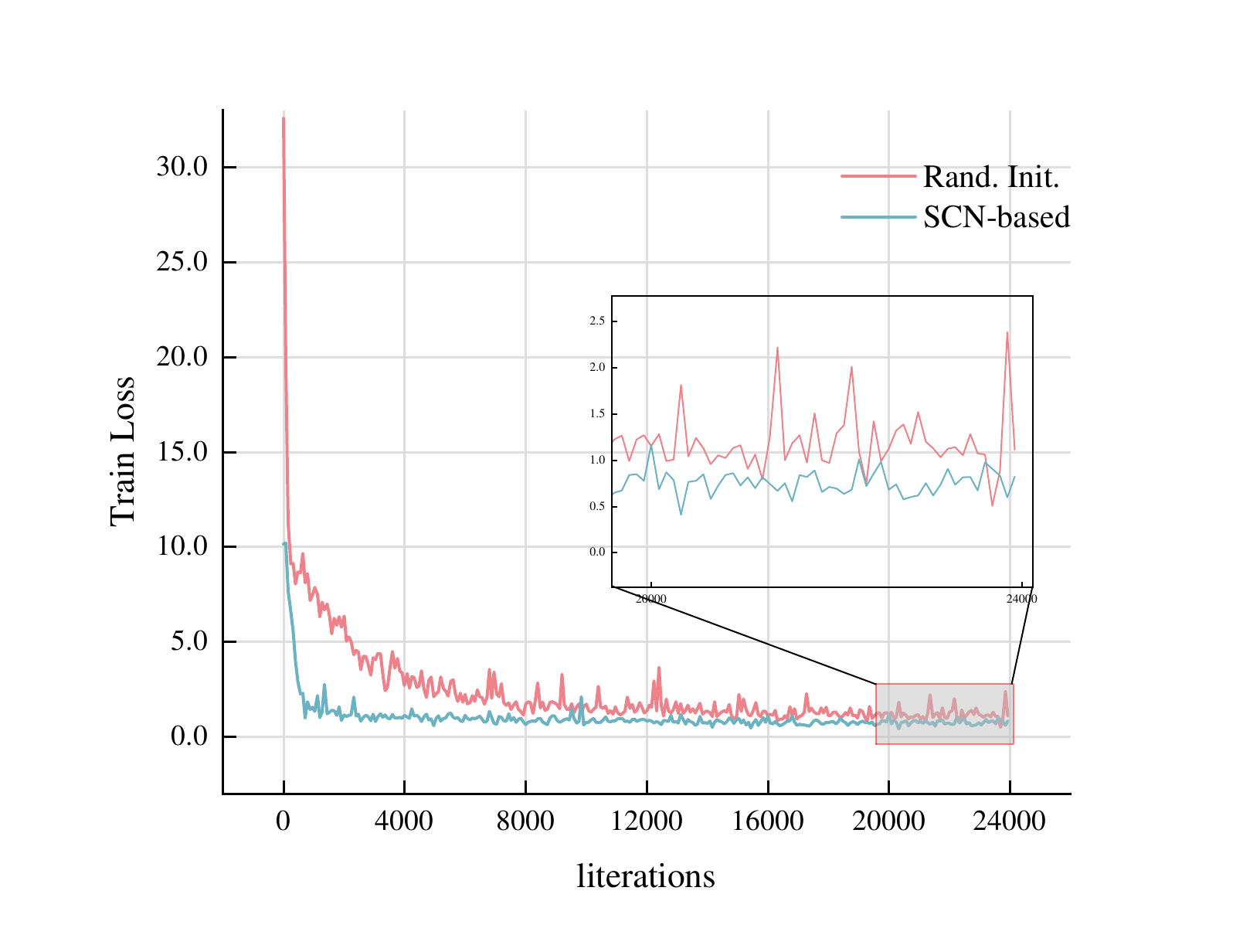}
  \caption{Training Loss Curves: Random Initialization vs. SCN-based Initialization}
  \label{fig-loss}
\end{figure}

Based on the data presented in Fig.~\ref{fig-loss}, the training loss curves of randomly initialized and SCN-based  weights under the 5\% label condition were compared, with the descent rate of the loss value serving as an approximate measure of the convergence rate. It is evident that SCN outperforms random initialization in terms of convergence, accelerating the early stages of convergence and ultimately achieving a lower final convergence loss.
\begin{table}[ht]
  \caption{Performance of SChanger on LEVIR-CD Dataset under Few-shot Learning. The F1 Scores (\%) are Highlighted.}
  \centering
  \begin{tabular}{l|c|c|c|c|c}
    \toprule
    \textbf{Sample Ratio} & \textbf{5\%} & \textbf{10\%} & \textbf{20\%} & \textbf{30\%} & \textbf{100\%} \\
    \midrule
    \textbf{Rand. Init.} & 86.76 & 89.54 & 90.74 &91.25 &92.54 \\
    \textbf{SCN-based} & 91.26 & 91.57 & 92.04 & 92.47 &92.87  \\
    \midrule
    \textbf{Improvement} & \textcolor{upgreen}{$\uparrow$ 4.50} & \textcolor{upgreen}{$\uparrow$ 2.03}  & \textcolor{upgreen}{$\uparrow$ 1.30} &\textcolor{upgreen}{$\uparrow$ 1.22}&\textcolor{upgreen}{$\uparrow$ 0.33}  \\
    \bottomrule
  \end{tabular}
  \label{tab:fewshot_performance}
\end{table}

\subsection{Transferability Evaluation}
A transferability evaluation of SChanger is conducted alongside various benchmark models. The models are initially trained using the LEVIR-CD dataset, and their performance is then assessed on the WHU-CD dataset. The accuracy metrics are shown in Table~\ref{tab:cross_domain_cd_comparison}.

Models trained on the LEVIR-CD dataset experience a significant drop in performance when applied to the WHU dataset. This decline can be attributed to differences in imaging conditions and scene characteristics between the two datasets. Among all the evaluated models, SChanger achieves the highest F1 score on the WHU dataset, demonstrating its superior transferability. Notably, SChanger-base shows a 20.43\% improvement over SGSLN/512, highlighting that SChanger is capable of extracting more generalized change features, rather than being constrained to a single scene.

\begin{table}[ht]
  \caption{Accuracy Comparison on the Cross-Domain CD from LEVIR-CD to WHU-CD.}
  \centering
  \begin{tabular}{l|ccc}
    \toprule
    \textbf{Method} & \textbf{P (\%)} & \textbf{R (\%)} & \textbf{F1 (\%)}  \\
    \midrule
    IFN~\cite{zhang2020deeply} & \textbf{78.92} & 21.73 & 34.08  \\
    BiT-18~\cite{chen2021remote} & 64.38 & 33.83 & 44.35  \\
    TransUNetCD~\cite{li2022transunetcd} & 72.28 & 38.47 & 50.21  \\
    FCCDN~\cite{chen2022fccdn} & 70.91 & 41.24 & 52.15  \\
    SGSLN/512~\cite{zhao2023exchanging} & 73.46 & \textbf{42.83} & \textbf{54.11}  \\
    \midrule
    SChanger-small & \textcolor{red}{\textbf{88.42}}   & \textcolor{softblue}{\textbf{52.01}} &\textcolor{softblue}{\textbf{65.49}} \\
    SChanger-base & \textcolor{softblue}{\textbf{87.59}}   & \textcolor{red}{\textbf{64.87}} & \textcolor{red}{\textbf{74.54}} \\
    \bottomrule
  \end{tabular}
  \label{tab:cross_domain_cd_comparison}
\end{table}

\subsection{Efficiency Test}
\begin{table*}[!t] 
  \caption{Efficiency Comparison of Different Methods: The Number of Flops and Throughput Computed Using a Tensor of Shape \(2 \times 3 \times 256 \times 256\).}\label{tab:efficacy}
  \centering
  \begin{tabular}{l|c|ccc|c}
    \toprule
    \textbf{Method} & \textbf{Year} & \textbf{Params. (M)} & \textbf{Flops (G)} & \textbf{Throughput (Sample/s)} & \textbf{LEVIR-CD F1 (\%)} \\
    \midrule
    FC-EF~\cite{daudt2018fully} & 2018 & 1.350 & 3.121  & 416.55 & 83.40  \\
    FC-Siam-Diff~\cite{daudt2018fully} & 2018 & 1.350 & 4.275  & 294.52 & 86.31 \\
    FC-Siam-Conc~\cite{daudt2018fully} & 2018 & 1.546 & 4.879  & 288.19 & 83.69 \\
    IFN~\cite{zhang2020deeply} & 2020 & 35.995 & 79.036 & 131.09 & 88.13 \\
    ChangeFormer~\cite{bandara2022transformer} & 2020 & 41.015 & 138.639 & 43.78 & 91.11 \\
    CGNet~\cite{han2023change} & 2023 & 38.989 & 87.683  & 189.07 & 92.01 \\
    SGSLN/512~\cite{zhao2023exchanging} & 2023 & 6.036  & 11.524 & 96.23 & 92.33 \\
    C2FNet~\cite{han2024c2f} & 2024 & 16.257 & 62.130  & 77.69 & 91.83 \\
    SRCNet~\cite{chen2024src} & 2024 & 5.195  & 8.373  & 86.57 & 92.24 \\
    SChanger-small (ours) & - & 0.607  & 6.242   & 85.91 & 92.45 \\
    SChanger-base (ours) & - & 2.370  & 18.275  & 73.93 & 92.87 \\
    \bottomrule
  \end{tabular}
\end{table*}
The performance of the SChanger model is compared with other CD models, focusing on the number of parameters, computational load, throughput, and F1 score measured on the LEVIR-CD dataset. Throughput is assessed by recording the start and end times using CUDA's Event function to ensure precise timing. To improve accuracy, the experiment is conducted three times, with each iteration processing 1,000 samples. The first 10 samples from each run are discarded to eliminate the impact of warmup time. Table~\ref{tab:efficacy} illustrates that SChanger-small, with just 0.607M parameters and 6.242G Flops, attains an F1 score of 92.45\%, highlighting its efficient design. On the other hand, SChanger-base, which has 2.370M parameters and 18.275G Flops, achieves a higher F1 score of 92.87\%. Remarkably, SChanger-small delivers a similar F1 score to SGSLN/512 while reducing the parameter count by nearly 90\% and Flops by 50\%, alongside comparable throughput, demonstrating its computational efficiency.
\section{Ablation Study}
\begin{table*}[!t]
  \caption{Ablation Study of Different Modules on the LEVIR-CD Dataset. \label{tab:ablation}}
  \centering
  \begin{tabular}{l|cccc|cc|ccc|c}
    \toprule
    \textbf{ID} & \textbf{MSFSH} & \textbf{VANM} & \textbf{SCAM} & \textbf{SCN} & \textbf{Params. (M)} & \textbf{Flops (G)} & \textbf{P (\%)} & \textbf{R (\%)} & \textbf{F1 (\%)}& \textbf{$\Delta$ F1 gains (\%)} \\
    \midrule
    1 & - & - & - & - & 1.857 & 14.505 & 92.87 & 91.09 & 91.97 &-\\
    2 & $\checkmark$ & - & - & - & 1.897 & 14.593 & 93.14 & 91.56 & 92.35&\textcolor{upgreen}{$\uparrow$ 0.38} \\
    3 & $\checkmark$ & $\checkmark$ & - & - & 2.304 & 18.381 & 91.77 & 92.97 & 92.37&\textcolor{upgreen}{$\uparrow$ 0.40} \\
    4 & $\checkmark$ & - & $\checkmark$ & - & 2.370 & 18.275 & 93.05 & 92.03 & 92.54&\textcolor{upgreen}{$\uparrow$ 0.57} \\
    5 & $\checkmark$ & - & $\checkmark$ & $\checkmark$ & 2.370 & 18.275 & 93.06 & 92.67 & 92.87&\textcolor{upgreen}{$\uparrow$ 0.90} \\
    \bottomrule
  \end{tabular}
\end{table*}
\begin{table*}[!t]
  \caption{Ablation Study of Different Strategies Used in SCN on the LEVIR-CD Dataset.}\label{tab:SCNcompare}
  \centering
  \begin{tabular}{l|cc|cc|ccc|c}  
    \toprule
    \textbf{ID} & \textbf{SAF} & \textbf{SFA} & \textbf{Param. (M)} & \textbf{Flops (G)} & \textbf{P (\%)} & \textbf{R (\%)} & \textbf{F1 (\%)} & \textbf{$\Delta$ F1 gains (\%)} \\
    \midrule
    1 & - & - & 2.267 & 18.261 & 92.42 & 91.11 & 91.76&- \\
    2 & - & $\checkmark$ & 2.304 & 18.381 & 92.55 & 91.74 & 92.15&\textcolor{upgreen}{$\uparrow$ 0.39} \\
    3 & $\checkmark$ & - & 2.333 & 18.155 & 92.69 & 91.86 & 92.27 &\textcolor{upgreen}{$\uparrow$ 0.51}\\
    4 & $\checkmark$ & $\checkmark$ & 2.370  & 18.275 & 93.06 & 92.67 & 92.87 &\textcolor{upgreen}{$\uparrow$ 1.11}\\
    \bottomrule
  \end{tabular}
\end{table*}
\subsection{Efficacy of MSFSH}
In the SChanger model, the MSFSH is crucial for detecting subtle changes by capturing multi-scale information 
and shortening the gradient backpropagation path.  As shown in Table~\ref{tab:ablation} (Experiment ID 1 and 2), 
the SChanger model with MSFSH outperforms the baseline, achieving a 0.38\% F1 score improvement, 
demonstrating the importance of MSFSH.
\subsection{Efficacy of SCAM}\label{subsec:eof}
In the SChanger model, SCAM enhances the receptive field and improves the interaction between bitemporal information streams. From a quantitative perspective, as shown in Table~\ref{tab:ablation}, VANM increases recall by 1.41\%, but reduces precision by 1.37\% (Experiment ID 2 and 3), likely due to its sensitivity to irrelevant features. However, incorporating a spatial consistency inductive bias (Experiment ID 3 and 4) improves both recall and precision while maintaining similar parameters and Flops. This results in a 0.57\% increase in the F1 score, underscoring the importance of feature interaction.

\begin{figure*}[!t]
  \centering
  \includegraphics[width=6.5in]{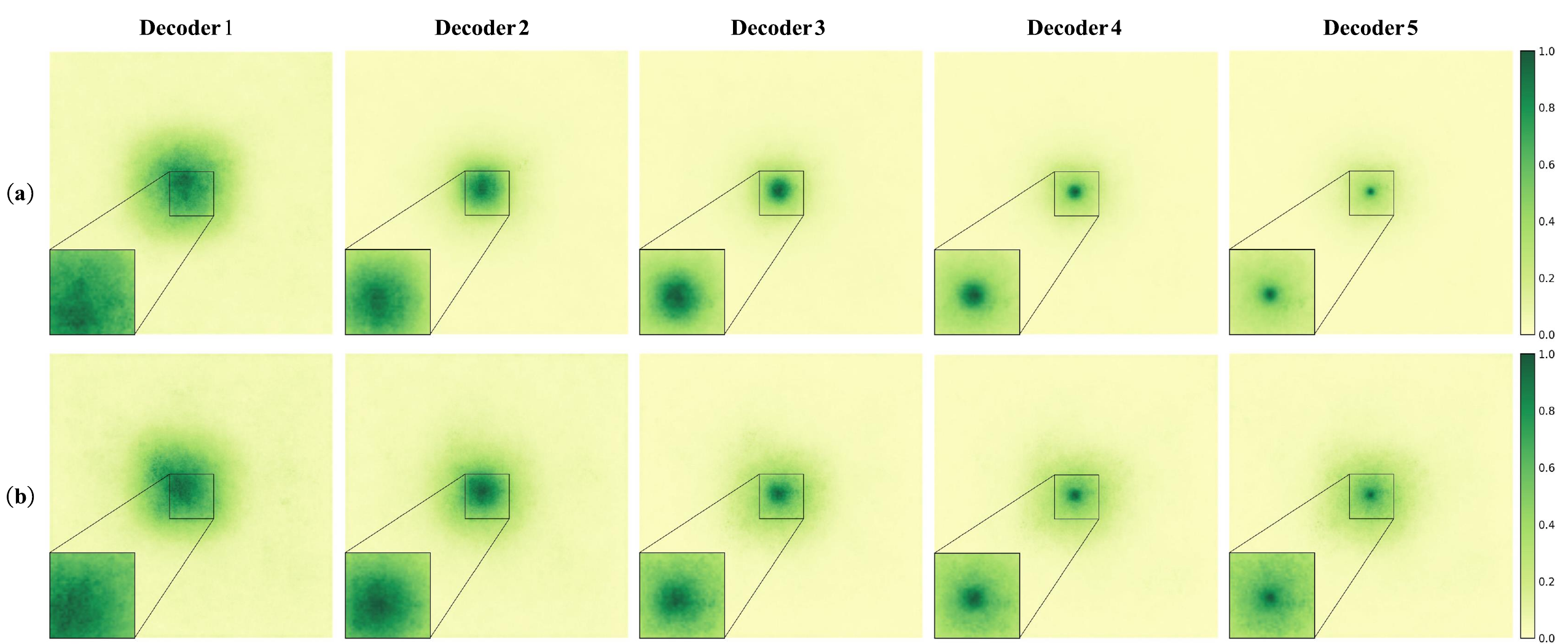}
  \caption{Effective Receptive Field of (a) w/o SCAM Model (Experiment ID 2), (b) w/ SCAM Model (Experiment ID 4).}
  \label{fig-erf}
\end{figure*}

From a qualitative perspective, the use of large-kernel convolutions leads to larger Effective Receptive Fields (ERFs)\cite{luo2016understanding}. We analyze Experiment ID 2 and 4, and Fig.~\ref{fig-erf} visually demonstrates how the ERFs evolve across different decoder stages in both the baseline model and the baseline with SCAM. Our observations are as follows: (1) As the decoder layers become deeper, the network focuses more on localized features, and the receptive field becomes more constrained. (2) SCAM significantly expands the ERF in the deeper decoder stages, which is crucial for dense prediction tasks in CD. These insights suggest that SCAM's use of large-kernel convolutions enhances the model's ability to capture long-range dependencies, leading to improved accuracy and feature representation.

\begin{figure*}[!t]
  \centering
  \includegraphics[width=6.5in]{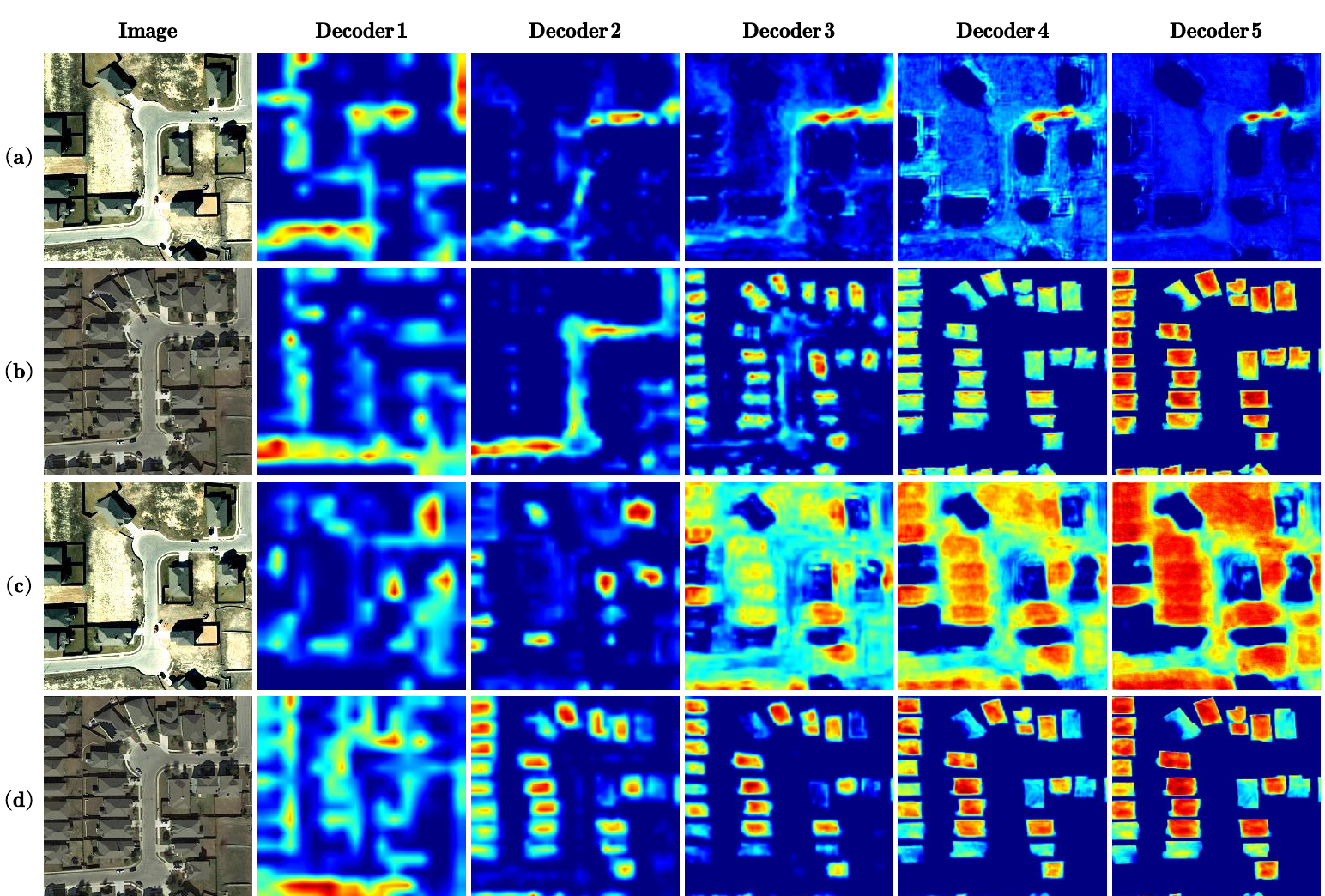}
  \caption{Grad-CAM visualization results (focusing on all pixels in the change category). The visualization comparison is made between the VANM (Experiment ID 3) and SCAM (Experiment ID 4). The panels show: (a) VANM for t1, (b) VANM for t2, (c) SCAM for t1, (d) SCAM for t2.}
  \label{fig-gradcam}
\end{figure*}

\begin{figure*}[!t]
  \centering
  \includegraphics[width=6in]{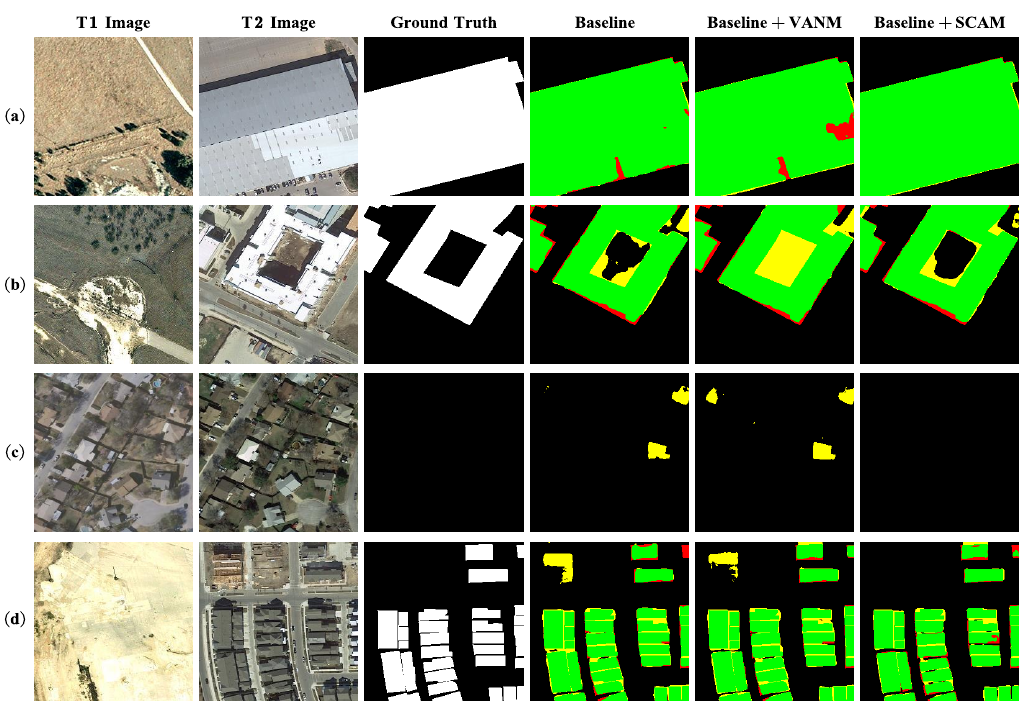}
  \caption{Error analysis for SChanger on LEVIR-CD. The rendered colors represent TP~(green), FP~(yellow), and FN~(red).}
  \label{fig-levircd_compare}
\end{figure*}

Furthermore, to better understand SCAM's effectiveness and superior performance, we use Grad-CAM~\cite{selvaraju2017grad} to analyze and compare Experiment ID 3 and 4. As illustrated in Fig.~\ref{fig-gradcam}, the heatmap analysis of the two decoder layers in the SChanger model highlights key differences. (1) The VANM variant uniformly focuses on buildings and their surroundings but struggles to distinguish between changed and unchanged structures, while SCAM accurately identifies and focuses on the changed buildings. (2) VANM overly emphasizes noisy road features, whereas SCAM filters out noise and concentrates on changes related to buildings. These findings emphasize SCAM's advantage in change detection, improving overall accuracy.
Fig.\ref{fig-levircd_compare} provides a visual comparison of SChanger variants on the LEVIR-CD dataset, showcasing their ability to detect and localize changes. For large-scale buildings, as shown in Fig.\ref{fig-levircd_compare}(a) and Fig.\ref{fig-levircd_compare}(b), the SCAM model excels at preserving fine details and edges, avoiding internal gaps. In cases where trees partially obstruct buildings, as in Fig.\ref{fig-levircd_compare}(c), SCAM effectively distinguishes between actual and obstructed instances, reducing false positives. In complex environments, as shown in Fig.~\ref{fig-levircd_compare}(d), SCAM demonstrates superior performance in localizing changed buildings, showcasing improved precision, robustness, and localization. These results confirm SChanger's effectiveness in accurately extracting change information.

\subsection{Efficacy of SCN}
The primary objective of training the SChanger model using the SCN strategy is to effectively integrate prior knowledge for instance extraction. To evaluate its impact, we load the pre-trained weights and conduct an ablation study focusing on the two key components of SCN, namely SFA and SAF. As shown in Table~\ref{tab:SCNcompare}, the results demonstrate that both the SFA and SAF modules significantly enhance the model's performance. When combined, these modules yield a 1.11\% improvement in the F1 score, underscoring their effectiveness in adapting pre-trained models for the change detection task. In comparison to models with random initialization, as shown in Table~\ref{tab:ablation} (Experiments ID 4 and 5), incorporating prior knowledge and fine-tuning results in a 0.64\% increase in recall and a 0.33\% improvement in the F1 score. These findings highlight the crucial role of prior knowledge in enhancing the model's ability to handle complex changes.
\subsection{Evaluation of Fusion Strategies}

To evaluate the TFM module, comparative experiments are conducted using different fusion strategies on the LEVIR-CD dataset. We ensure fairness by keeping the parameter counts and computational complexity similar across all strategies. We apply the SCN strategy and compare four fusion methods: direct addition, absolute subtraction, TFM-BN (using BN), and TFM-LN (using LN). As shown in Table~\ref{tab:levir_cd}, compared to the widely-used baseline of absolute subtraction, TFM-LN improves recall by 0.69\% and the F1 score by 0.14\%. Additionally, compared to TFM-BN, TFM-LN improves recall by 1.38\% and the F1 score by 0.47\%. These results confirm that LN preserves temporal information and enhances overall performance in CD tasks.

\begin{table*}[!t]
  \caption{Comparison of Different Fusion Strategies on the LEVIR-CD Dataset. The Best Values are Highlighted in Bold.\label{tab:levir_cd}}
  \centering
  \begin{tabular}{l|cc|ccc}
    \toprule
    \textbf{Fusion Strategy} & \textbf{Params. (M)} & \textbf{Flops (G)} & \textbf{P (\%)} & \textbf{R (\%)} & \textbf{F1 (\%)} \\
    \midrule
    Addition            & 2.265  & 17.860 & 93.16  & 92.00  & 92.57  \\
    Absolute Subtraction & 2.265  & 17.860 & 93.50  & 91.98  & 92.73  \\
    TFM-BN              & 2.370  & 18.303 & \textbf{93.53}  & 91.29  & 92.40  \\
    TFM-LN              & 2.370  & 18.275 & 93.06  & \textbf{92.67}  & \textbf{92.87}  \\
    \bottomrule
  \end{tabular}
\end{table*}

\section{Discussion}
\subsection{Negative Transfer}
  In an effort to further enhance the generalization capability of SChanger, the SPNet model is pre-trained on a larger dataset. The aim is to improve its performance across diverse tasks. However, the results are unexpectedly contrary to expectations. The model is pre-trained on the WHU-Mix dataset~\cite{luo2023diverse}, which includes regions from the Kitsap, Tyrol, and Vienna areas of the IAILD dataset, as well as a diverse set of regions from five continents. The pre-training configuration is consistent across all experiments.
  \begin{table}[ht]
    \centering
    \caption{Cross-Domain Evaluation of SPNet and Fine-Tuned Scores for SCN Pre-trained Model on Different Datasets}
    \begin{tabular}{l|c|ccc}
    \toprule
    \textbf{Init. Method} & \textbf{MA} &\textbf{LEVIR-CD} & \textbf{SYSU-CD} & \textbf{WHU-CD} \\
    \midrule
    Rand. Init. &-& 92.54 & 83.95 &  92.90\\
    IAILD &27.17& 92.87 & 84.17 & 93.20 \\
    WHU-Mix& 59.29& 92.56  & 84.05  & 92.26  \\
    \midrule
    Change& \textcolor{upgreen}{$\uparrow$ 32.13}& \textcolor{red}{$\downarrow$ 0.31} & \textcolor{red}{$\downarrow$ 0.12} &  \textcolor{red}{$\downarrow$ 0.94} \\
    \bottomrule
    \end{tabular}
  
    \label{tab:init_scores}
    \end{table}
  To evaluate whether the model successfully learns key building features, a Cross-Domain test is conducted using the Massachusetts Buildings Dataset (MA)~\cite{MnihThesis}. The results show that the pre-trained model has indeed acquired useful knowledge for building extraction. As illustrated in Table~\ref{tab:init_scores}, the model pre-trained on the WHU-Mix dataset outperforms the IAILD-pretrained model by 32.13\% in the Cross-Domain experiment on the MA dataset, highlighting the model's ability to generalize and extract key building features.
  
  However, when fine-tuned using the SCN method on CD datasets, the performance of the WHU-Mix pre-trained model is found to be inferior to that of the IAILD pre-trained model across all three datasets: LEVIR-CD, SYSU-CD, and WHU-CD. Specifically, the performance decreases by 0.31\%, 0.12\%, and 0.94\%, respectively. Furthermore, on the WHU-CD dataset, the model pre-trained on WHU-Mix even performs worse than the randomly initialized model. This unexpected discrepancy in performance prompts the occurrence of negative transfer~\cite{wang2019characterizing}. Given that the IAILD pre-trained model outperforms the randomly initialized model, it is unlikely that the building extraction task itself hinders the CD task. One possible explanation is that the WHU-Mix dataset contains a significant amount of irrelevant data, which may require data cleaning or the application of data generation techniques to reduce the influence of low-quality samples and increase the number of high-quality samples.
\subsection{Expectations and Limitations}
Bitemporal change detection is a foundational task in the broader field of CD, focused on identifying and extracting areas that have undergone changes by comparing dual-temporal image data. Various methods have been proposed to address this task, with dual-branch networks emerging as a widely adopted approach. In this paper, we introduce SCN, a novel method that enhances bitemporal change detection performance by leveraging more data from single-temporal images. The SChanger model trained using SCN demonstrates strong performance across multiple binary change detection and object change detection tasks. However, with the continuous advancements in Earth Observation technologies, the availability of multi-temporal image data has increased significantly. As a result, processing visual data across multiple time points has become an increasingly essential challenge for more complex CD tasks.

While dual-branch networks attempt to iteratively process multi-temporal images by taking two images at a time, this approach has a significant limitation. Specifically, for n images, the encoder stage feature extraction is wasted n-2 times, leading to inefficiencies in processing and underutilization of available data. Additionally, binary change detection methods commonly used in these networks do not provide clarity on the categories of changes, which limits the model's ability to extract meaningful and actionable information from multi-temporal datasets. 

\section{Conclusion}
We propose the SChanger model for CD, integrating SCAM, TFM, LFEM, and MSFSH, trained with the SCN strategy. SCAM incorporates spatial consistency to enhance bitemporal fusion. TFM employs LN to stabilize training by normalizing features while preserving temporal information. The SCN strategy utilizes fully pretrained weights from single-temporal tasks, facilitating accurate change identification. These components collectively contribute to the superior performance of SChanger in CD tasks.

Experimental results indicate that SChanger surpasses all benchmark models across multiple datasets, demonstrating notable efficiency, robustness, and scalability. These findings suggest that SChanger has the potential to serve as a benchmark for CD tasks. Future work, from the data perspective, will focus on exploring semi-supervised CD, as well as zero-shot and few-shot CD techniques, to further enhance the model's adaptability and performance in scenarios with limited labeled data. From the model perspective, the focus will be on extending the model's capability to handle multi-temporal images.

\section*{Acknowledgments}
This work was supported by the National Natural Science Foundation of China (No. 42376180), the Key R\&D Project of Hainan Province (ZDYF2023SHFZ097) and the Hohai University Undergraduate Innovation and  Entrepreneurship Training Program Funding~(202410294199Y).


\bibliography{paper.bib}
%
\bibliographystyle{IEEEtran}

\newpage

\vspace{11pt}

\begin{IEEEbiography}[{\includegraphics[width=1in,height=1.25in,clip,keepaspectratio]{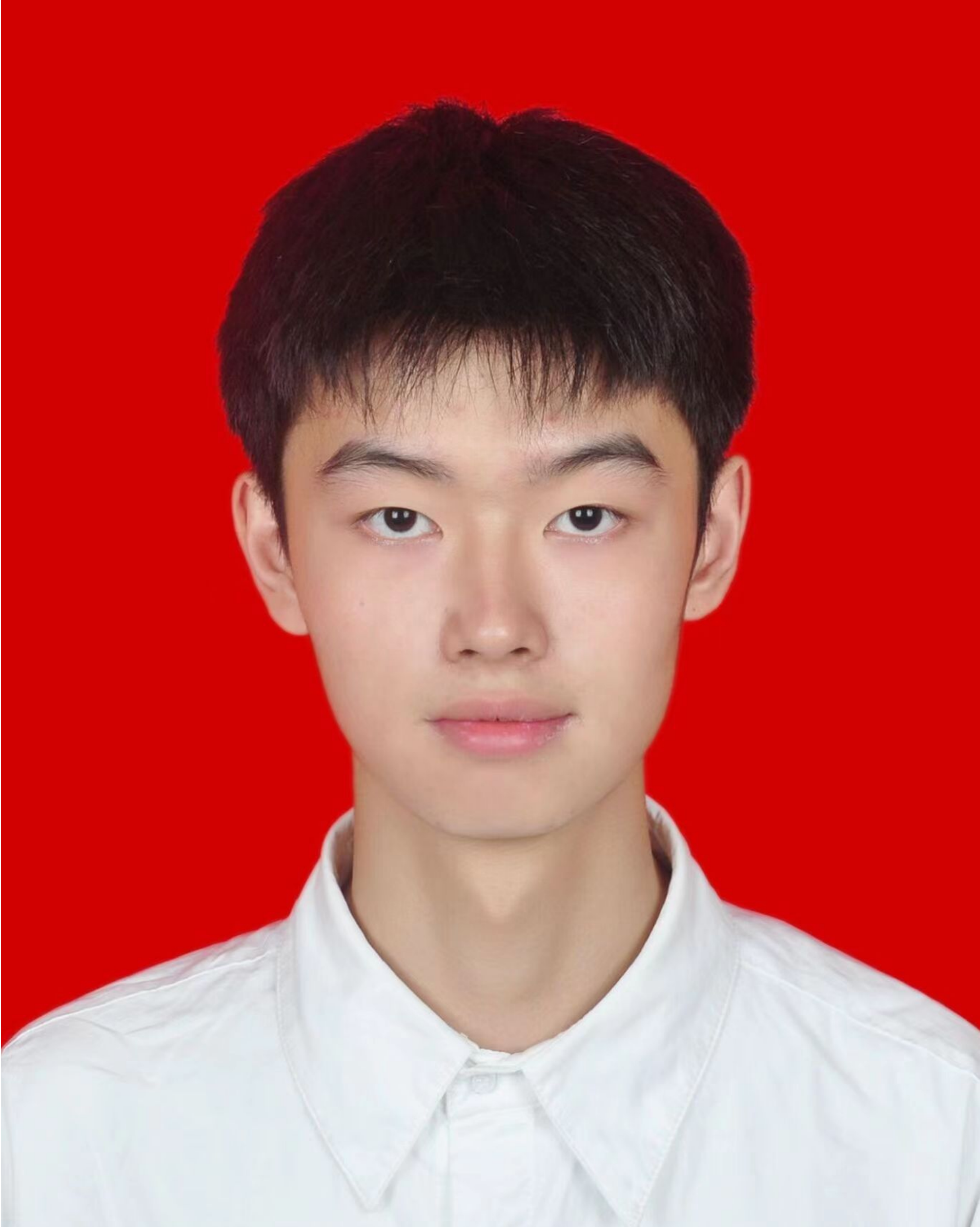}}]{Ziyu Zhou}
  is currently working toward the B.S. degree in School of Earth Sciences and Engineering, Hohai University, Nanjing, China. His research interests include deep learning in remote sensing, change detection and generative models.
\end{IEEEbiography}

\begin{IEEEbiography}[{\includegraphics[width=1in,height=1.25in,clip,keepaspectratio]{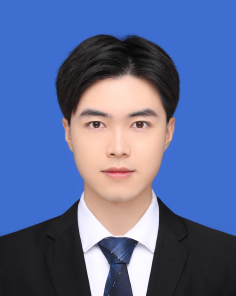}}]{Keyan Hu}
  received the B.S. degree in Surveying and Mapping Engineering from the School of Earth Sciences and Engineering, Hohai University, Nanjing, China, in 2024. He is currently working toward the M.S. degree in Photogrammetry and Remote Sensing with the School of Geosciences and Info-Physics, Central South University, Changsha, China. His research interests include deep learning applications in remote sensing image semantic segmentation, multimodal learning, and generative models.
\end{IEEEbiography}

\begin{IEEEbiography}[{\includegraphics[width=1in,height=1.25in,clip,keepaspectratio]{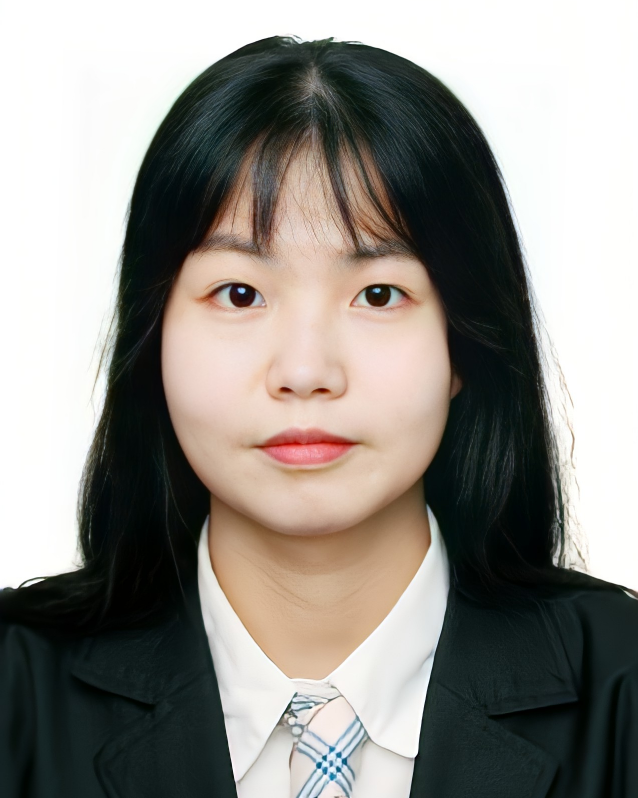}}]{Yutian Fang}
  is currently working toward the B.S. degree in School of Earth Sciences and Engineering, Hohai University, Nanjing, China. Her research interests include deep learning applications in remote sensing image semantic segmentation, multimodal learning, and generative models.
\end{IEEEbiography}

\begin{IEEEbiography}[{\includegraphics[width=1in,height=1.25in,clip,keepaspectratio]{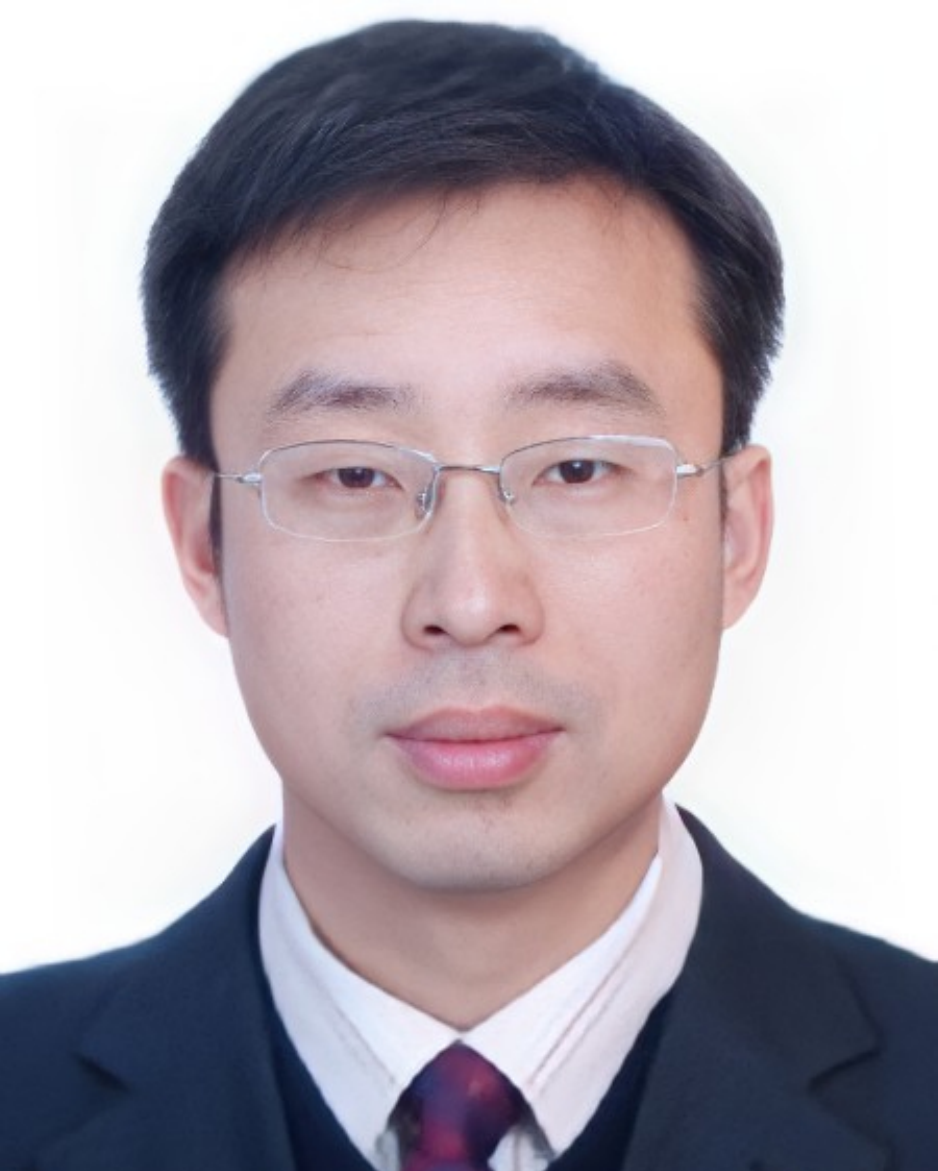}}]{Xiaoping Rui}
  received PhD degree in Cartography and Geographic Information System from the Graduate University of Chinese Academy of Sciences, Beijing,China, in 2004. He is currently a full professor with the School of Earth Sciences and Engineering, Hohai
  University. His research interests include geographical big data mining, 3D visualization of spatial data, and remote sensing image understanding.
\end{IEEEbiography}

\vspace{11pt}

\vfill

\end{document}